\documentclass[11pt,a4paper]{article}
\usepackage[hyperref]{emnlp2020}
\usepackage{times}
\usepackage{array}
\usepackage{latexsym}
\usepackage{subcaption}
\usepackage{booktabs}
\usepackage{graphicx}
\usepackage{placeins}
\usepackage{amssymb}

\usepackage{graphics}
\usepackage[normalem]{ulem}
\usepackage{stmaryrd}
\usepackage{tikz}
\usepackage{tikz-dependency}

\usepackage{xr}
\makeatletter
\newcommand*{\secl}[0]{{\sc SErCl}}
\newcommand*{\addFileDependency}[1]{
	\typeout{(#1)}
	\@addtofilelist{#1}
	\IfFileExists{#1}{}{\typeout{No file #1.}}
}
\makeatother

\usepackage{url}

\aclfinalcopy 

\usepackage[utf8]{inputenc}
\usepackage{microtype}
\usepackage{relsize}

\usepackage[font=10pt,labelfont=bf,skip=5pt]{caption}

\usepackage{xcolor}
\usepackage{multirow}

\newcommand{\ra}{\(\shortrightarrow\)}
\newcommand{\com}[1]{}
\title{Classifying Syntactic Errors in Learner Language}

\author{ 
	Leshem Choshen$^{*}$\\
	Department of Computer Science\\
	Hebrew University of Jerusalem\\
	{\tt\small leshem.choshen@mail.huji.ac.il}\\
	\And
	Dmitry Nikolaev\Thanks{ First two authors contributed equally.}\\
	Department of Linguistics \\
	Stockholm University \\
	{\tt\small dnikolaev@fastmail.com}\\
	\AND Yevgeni Berzak\\
	BCS \\
	MIT\\
	{\tt\small berzak@mit.edu}\\
	\And Omri Abend \\
	Department of Computer Science\\
	Hebrew University of Jerusalem\\
	{\tt\small omri.abend@mail.huji.ac.il}\\
}

\date{}

\begin{document}
	\maketitle
	
	\begin{abstract}
		We present a method for classifying syntactic errors in learner language, namely errors whose correction alters the morphosyntactic structure of a sentence.
		The methodology builds on the established Universal Dependencies syntactic representation scheme, and provides complementary information to other error-classification systems. 
		Unlike existing error classification methods, our method is applicable across languages, which we showcase by producing a detailed picture of syntactic errors in learner English and learner Russian. We further demonstrate the utility of the methodology for analyzing the outputs of leading Grammatical Error Correction (GEC) systems.
	\end{abstract}
	
	\section{Introduction}
	\com{
		Remember to recheck:
		intro
		each section place is clear
		reiterate paragraph
		
		structure:
		taxonomy explained
		validations of our taxonomy+classification (e.g. parsers work well)
		comparison to other taxonomies /classifications (manual and ERRANt) 
		proof of usefulness
		various kinds of analysis this allows (further proof of usefulness)
		qualitative results and discussion
		related work
		conclusion
	}
	Taxonomies of grammatical errors are important for linguistic and computational analysis of learner language, as well as for Grammatical Error Correction (GEC) systems.\footnote{Code can be found \href{https://github.com/borgr/GEC_UD_divergences}{in github repo GEC\_UD\_divergences}. Matrices directly mentioned are included in the appendix.}
	Such taxonomies divide the complex space of errors into meaningful categories and enable characterizing their distribution in learner productions. This information can be beneficial for GEC: it can support the development of systems that focus on specific error types, serve as a form of inductive bias (for example, by regularizing the system's output to have a desired distribution over correction types), and guide data augmentation and data filtering by controlling the distribution of error types. Error taxonomies can also improve the interpretability of system outputs for error analysis and learner feedback.
	

	\begin{figure}[t]
		\begin{small}
			\begin{dependency}
				\centering
				\hspace{-0.15cm}\begin{deptext}
					... \& if \& you \& like \& a \& subject \& you \& 'll \& study \& it \& {\bf easier}$_{ADJ}$ \\
				\end{deptext}
				\depedge[edge height= .3cm, edge style={red!60!black,ultra thick}]{9}{11}{xcomp}
				\node (source) [above of = \wordref{1}{2}, yshift = -0.5cm] {\textbf{Source:}};
			\end{dependency}
			\begin{dependency}
				\centering
				\hspace{-0.15cm}\begin{deptext}
					... \& if \&  you \& like \& a \& subject \& you \& 'll \& study \& it \& {\bf more} \& {\bf easily}$_{ADV}$ \\
				\end{deptext}
				\depedge[edge height= .6cm, edge style={red!60!black,ultra thick}]{9}{12}{advmod}
				\depedge[edge height= .2cm, edge style={red!60!black,ultra thick}]{12}{11}{advmod}
				\node (reference) [above of = \wordref{1}{3},xshift=-0.2cm, yshift = -0.5cm] {\textbf{Reference:}};
			\end{dependency}
			\caption{\small\label{fig:example} Example of an edit of type ADJ~\ra~ADV in POS terms and \texttt{xcomp}~\ra~\texttt{advmod} in edge-label terms. Corresponding spans are boldfaced. }
		\end{small}
		\vspace{-0.5cm}
	\end{figure}

	A number of annotation efforts for learner language developed error taxonomies \citep{nicholls2003cambridge,dahlmeier2013building}, and statistical classifiers into such taxonomies, notably ERRANT \citep{bryant2017automatic}. Taking error types into consideration in learning has also been shown to improve GEC performance \citep[][{cf.  \S\ref{sec:related}}]{kantor2019learning}.
	However, most existing taxonomies are fairly coarse-grained and language specific, and do not produce meaningful types for a large proportion of the errors. For example, 25\% of the errors in the standard NUCLE corpus \citep{dahlmeier2013building} are mapped to the residual category OTHER (see \S\ref{sec:errant}).
	
	We propose \secl, a taxonomy of {\it \textbf{S}yntactic \textbf{Er}rors} (SEs) and an automatic \textbf{Cl}assification. Inspired by a longstanding tradition in Machine Translation (MT) which analyses divergences between source and translated texts based on syntactic structure \citep{dorr1994machine, nikolaev2020fine}, \secl\ is based on divergences between ungrammatical sentences and their corrections.
	We define SEs as errors whose correction involves changing morphological features, POS labels or the syntactic structure labels. \secl\ takes as input edits, i.e., grammatically incorrect text spans and their corrections, and compares their labels. For example, the error in Fig. \ref{fig:example} is an adjective replaced with an adverb (ADJ\ra ADV) in POS terms, and an \texttt{xcomp}\ra\texttt{advmod} in edge-label terms. Thus, SEs are defined by changes in form, rather than by the principles governing the choice of a correct form.
	
	\secl\ is the first taxonomy derived from a syntactic representation framework, and it uses the Universal Dependencies formalism \citep[UD;][]{nivre2016universal}. 
	This approach provides three major advantages over prior learner error taxonomies. First, the \secl\ taxonomy is derived automatically from UD annotations, circumventing the need for constructing ad-hoc manually defined error categories. Second, using the UD formalism makes the method applicable across languages, allowing for consistent analyses and comparisons of learner errors across different languages within one unified framework. Third, \secl\ is compatible with standard representations and tools in NLP. 
	
	Further, the UD based approach to error classification can yield finer distinctions compared to existing schemes. For example, it divides the commonly used class of adposition errors into errors in the use of prepositions as nominal modifiers (e.g., ``a mention \sout{to} {\bf of} previous work''), and the use of prepositions in prepositional objects or adjuncts (e.g., ``referring \sout{for} {\bf to} previous work''). 
	POS tags alone cannot distinguish them, but the UD trees expose this distinction straightforwardly. UD can also help classify agreement and case-assignment errors thanks to its morphological-feature layer containing information about case, number, gender, and other features relevant for inflection.

	We validate \secl's reliability by showing (1) SEs based on automatic parses are similar to ones based on manual parses.  (\S\ref{sec:auto_parse}); (2) \secl\ types map well to NUCLE's manually curated taxonomy (\S\ref{sec:nucle}); (3) \secl\ is complementary to the standard type classifier ERRANT: 60\% of the errors not classified by ERRANT are classified by \secl.
	
	We demonstrate \secl's unique features, notably cross-linguistic applicability, by analyzing SE distributions in available corpora for learner English (\S\ref{sec:corpus_study_tle}) and learner Russian (\S\ref{sec:multi_language}).
	
	Finally, we find in GEC systems (1) certain SEs are harder to correct (2) SEs are harder than non-SEs (c.f. \ref{sec:system_outputs}) (3) the granular types can help devising rules to improve products (e.g. Grammarly, \S\ref{sec:uncommon}).

	\section{Methodology}\label{sec:methodology} 
	
	
	
	This section defines our taxonomy of SEs and how \secl\ classifies into it. Given a parsed learner sentence and its correction, and given an edit $e=(e_s,e_c)$, i.e., a sub-string of the source sentence $e_s$ that contains a grammatical error and its reference correction $e_c$, we define its class in the following way. We select a representative token in $e_s$ and in $e_c$. Specifically, each sub-string defines a sub-forest of the dependency parse, and the representative is taken to be the node closest to the root.\footnote{We select the leftmost token to break ties (3.5\% of TLE SEs).} The rationale for this decision is that UD treats grammatical markers as dependents of content words. 
	Therefore, in most cases the semantic and syntactic heads correspond to one another, even if lexical items are changed. For example, in \textit{\textbf{went}}\ra\textit{was \textbf{walking}}, the semantic and syntactic head of the target has the lemma \textit{walk} and not \textit{be}.\footnote{In order to investigate the type correspondences of other tokens in the sub-strings, we may extract dependents of the representative nodes and compare their labels. This is an avenue for future work.}
	
	We define an SE as an edit where the two representative's labels do not match. The SE type is defined as the ordered pair of labels with the source label going first. Special cases of SEs are additions and deletions, i.e., edits in which the source or target span is empty.
	
	This definition of SEs is straightforward to implement and requires no further annotation on top of the edits and parses, but it leaves out cases where the representative tokens retain their labels (e.g., agreement errors or inappropriate determiners), although these distinctions can be made in some cases based on UD's morphological features. For practical use, one can annotate all these non-SE errors by the feature that is retained (e.g. Plural Noun errors, if POS tag and morphological features are used). Given a corpus, a confusion matrix could be extracted, where the diagonal counts the non-SEs. 
	
	We focus in this work on universal POS-tag pairs, which are sufficient to classify and explain the majority of SEs in English. Dependency labels are analyzed as well, although we find that edge-label-based types and POS-based types are strongly correlated (\S\ref{sec:corpus_study_tle}).
	We also explore the use of morphological features, and apply it to Russian that has a rich morphology (\S \ref{sec:multi_language}).

	\begin{table}[t]
		{\small
			\begin{tabular}{@{}ll@{}}
				\multicolumn{1}{c}{{Acronym + Ref.}}   & \multicolumn{1}{c}{{Notes}} \\ 
				\toprule
				TLE \citep{berzak2016universal}     & Manual parses \\
				\midrule
				NUCLE           & \multirow{2}{*}{Standard GEC benchmark}  \\           
				\citep{dahlmeier2013building} & \\
				\midrule
				Lang8          & \multirow{2}{*}{No error classes} \\
				\cite{Mizumoto2012TheEO} & \\
				\midrule
				\multirow{2}{*}{W\&I       \citep{bryant-etal-2019-bea}}    & Varied proficiency levels\\ 
				&  ERRANT classes  \\
				\midrule
				RULEC       & \multirow{2}{*}{Learner Russian}             \\                            
				\citep{rozovskaya2019grammar} & \\
				\bottomrule
			\end{tabular}
		}
		\caption{Datasets used in this work.\label{tab:datasets}}
		\vspace{-0.5cm}
	\end{table}
	
	
	
	
	\section{Reliability}
	\subsection{Reliance on Automatic Parses} \label{sec:auto_parse}
	
	\secl\ relies on syntactic trees. Manual annotation is currently only available for the TLE corpus \cite[][all datasets addressed in the paper are summarized in Table \ref{tab:datasets}]{berzak2016universal}, which includes POS and dependency relations, but no morphological features. Hence, we assess the outcomes of using a UD parser instead. We use UDPiPE \cite{Straka2016UDPipeTP} as our parser of choice. It is among the top-scoring parsers in CoNLL 2018 shared task \citep{zeman2018conll} for both English and Russian, the languages we consider in this paper.\footnote{A number of works designed parsers with learner language specifically in mind. However, as such parsers exist only for learner English, we use UDPipe for uniformity.}
	
	We begin by comparing the distribution of SEs in automatically parsed TLE and manual TLE. We focus the discussion on POS-based SEs for concisness; the full distributions of SE types, both edge-label-based and POS-based are found in Appendix \S\ref{sec:matrices_TLE}. When normalising by the number of tokens per POS, class frequencies are quite close to those obtained by manual edits (0.4\% absolute change on average and Pearson correlation of $r=0.998$). This is also the case when normalising by the amount of SEs per POS (0.05\% change). The results suggest that the use of a parser does not qualitatively change the distribution of SEs, and that current UD parsing technologies are mature enough to be used for extracting SEs. 
	
	While trends are similar with manual and automatic parses, perhaps unsurprisingly, more SEs are found when using automatic parses. This is particularly clear for the ``other'' tag ``X'' and for interjections. Symbols are the only category where we find less SEs. We ignore these non-lexical tags in our analysis, suspecting that this is a weakness of the parser.
	Finding the parsing reliable, we move to compare \secl\ to existing approaches.
	
	
	\begin{table}[ht]
		\centering
		\begin{small}
			\begin{tabular}{ll}
				\begin{tabular}{@{}lll@{}}
					\toprule
					Source & Target & \# \\ 
					POS & POS  &  \\ 
					\midrule
					NOUN & VERB & 51 \\
					NOUN & ADJ & 50 \\
					ADJ & NOUN & 49 \\
					VERB & NOUN & 46 \\
					VERB & ADJ & 37 \\
					DET & PRON & 34 \\
					PRON & DET & 32 \\ \bottomrule
				\end{tabular}
				&
				\begin{tabular}{@{}lll@{}}
					\toprule
					\begin{tabular}[c]{@{}l@{}}Source \\ label\end{tabular} & \begin{tabular}[c]{@{}l@{}}Target \\ label\end{tabular} & \# \\ \midrule
					compound & amod & 32 \\
					cop & aux & 32 \\
					xcomp & ccomp & 32 \\
					obl & obj & 26 \\
					obl & advmod & 25 \\
					det & nmod:poss & 25 \\
					advmod & obl & 24 \\ \bottomrule
				\end{tabular}
			\end{tabular}
			\caption{Most prevalent types of SEs involving replacement in the TLE in terms of POS tags (left) and edge labels (right). Numbers are absolute counts. See \S\ref{ssec:replacement_examples} in the supplementary material for example sentences.} \label{table:pos_label_replacements}
		\end{small}
		\vspace{-0.3cm}
	\end{table}

	
	\subsection{Comparing to Manually Typed Edits}\label{sec:nucle}
	
	Unlike many NLP tasks, this work does not aim to mimic human behavior. Still, there is sense in comparing \secl\ to a manually annotated taxonomy. We compare NUCLE annotated train errors and \secl's (confusion matrix in appendix Table \ref{tab:m2_stx}). We ignore relocation errors as edits lack the necessary information to discern relocation from deletion.
	
	SE types are generally contained within a single NUCLE error type.  Indeed, on average 62\% of the instances of a given SE type are contained in the maximally overlapping NUCLE category, i.e., when assigning each SE a NUCLE category most of the SE's instances are NUCLE's category instances as well. 82\% of the instances on average belong to one of the three maximally overlapping NUCLE categories. CCONJ\ra ADV, for example, is almost solely (95\%) mapped to "transition" error type, addressing linking and phrase errors. This shows that SE types contain much of the information conveyed by NUCLE types.
	Qualitatively, \secl\ has more categories and splits NUCLE types to meaningful sub-types. It is thus usually more informative. For example, the "article or determiner" NUCLE type is split to insertions and deletions of determiners in addition to other SEs (mostly from or to determiner). 
	
	\subsection{Comparing to the Automatic ERRANT}\label{sec:errant}
	
	This section studies the relation between \secl's predictions and those of ERRANT. For comparability, we apply \secl\ to the edit spans produced by ERRANT. For brevity, we focus on POS-based SEs.
	
	ERRANT \citep{bryant2017automatic} is essentially the only classifier in use today, and is therefore a natural point of comparison. ERRANT taxonomy is coarse-grained. It assumes for the most part that POS tags are not altered in corrections, classifying many errors by their POS tag (e.g. adverb error). Consequently, ERRANT covers mostly spelling and word-form errors.
	
	We note three important differences between \secl\ and ERRANT.
	First, being based on UD, \secl\ is applicable across languages (see \S\ref{sec:multi_language}), while ERRANT requires new rules or other modifications per language \citep{boyd2018using}. Second, relying on an established framework with broad usability accords validity to \secl's taxonomy, which is otherwise hard to validate \citep{bryant2017automatic}. Last, ERRANT classifies most 
	SEs as {\sc Other}. \secl\ therefore complements ERRANT and is able to classify what ERRANT leaves unclassified. 
	
	Empirically, we find that ERRANT does not meaningfully classify a large portion of the errors: about 25\% of ERRANT's predictions fall into the residual category Other in NUCLE and Lang8, and about 15\% of them in W\&I and TLE. We analyze which of those Other edits are SEs, finding most of them are. 
	In W\&I, of the 842 errors classified as {OTHER}, only 338 errors (40.1\%) are cases where the POS remains unaltered, while the remaining 504 errors (59.9\%) are POS-based SEs. 
	The effective number of SE types that {\sc Other} classifies into is 80.6, i.e., an entropy of 4.4 nets of the POS-based type distribution in edits classified as {OTHER}.
	
	As for SEs not classified as OTHER, our manual analysis reveals that there too \secl\ provides complementary information to ERRANT. Of the remaining errors, 620 are POS-based SEs, while 3211 are not (19.3\%). 
	Leaving out errors that involve punctuation leaves us with 522 SEs in W\&I. Of those, the most common class is ``morphological inflection'' (indicating that the correction and the source share a lemma). On it, \secl\ provides additional information, e.g., that the most frequent morphological inflection SE is NOUN\ra ADJ (31\% of the cases), while the reverse direction is much rarer (7\%).
	The second most common type, spelling, proved to be challenging for the parser and ERRANT is hence more informative for those. This is also the case for word-order errors. While verb errors are only the third most common, together with its subcategories, such as {\sc verb:form}, they account for 131 SEs. These might benefit from the SE categorization of common cases (e.g., VERB$\leftrightarrow$AUX errors suggest an error in the syntactic structure, unlike non-SE errors that usually involve lexical selection). Similarly, the 56 orthography errors could benefit from subcategorization of the common errors. For example, NOUN\ra PROPN is a common orthography error by ERRANT; ERRANT's type thus does not specify that it is a proper noun lacking capitalization. The other cases are either similar in spirit and can benefit from categorization of frequently appearing SEs, or cases where the POS tagging of the source and target disagree, either due to the UD guidelines or to parser inconsistency (e.g., the source parse may consider a word a particle, while the target parse considers it an adposition).
	
	To conclude, about 60\% of the errors classified as OTHER by ERRANT receive a POS-based SE class. SE classification further provides non-trivial information in many instances of other ERRANT categories. Together, these demonstrate that \secl\ provides value beyond ERRANT's classification, even where only English is considered.
	

	\vspace{-.1cm} 
	\section{Cross-linguistic Corpus Studies} \label{sec:datasets}
	
	In this section, we apply \secl\ to available datasets, comparing between different English datasets, originally annotated in different taxonomies, and between English and Russian datasets.
	
	\subsection{English}\label{sec:corpus_study_tle}
	
	We analyze the English datasets and learner language characteristics through SEs. We start with TLE, which provides manual UD and edit annotation. Our analysis is based upon the available tokenization and edit annotations. To avoid double-counting, we merge overlapping edits to form a set of non-overlapping ones.
	After removing some noise in the XML markup, we extract 4584 SEs. 
	Of those, 2042 are additions, 1048 are deletions, and 1495 are replacements. In 657 cases of replacement, both the POS and  edge label are changed; in 306 cases, only the POS is changed; in 532 cases, only the edge-label is changed. 
	
\begin{figure*}[t]
		\centering
		
		\hspace*{-.5cm} \includegraphics[width=1.05\textwidth]{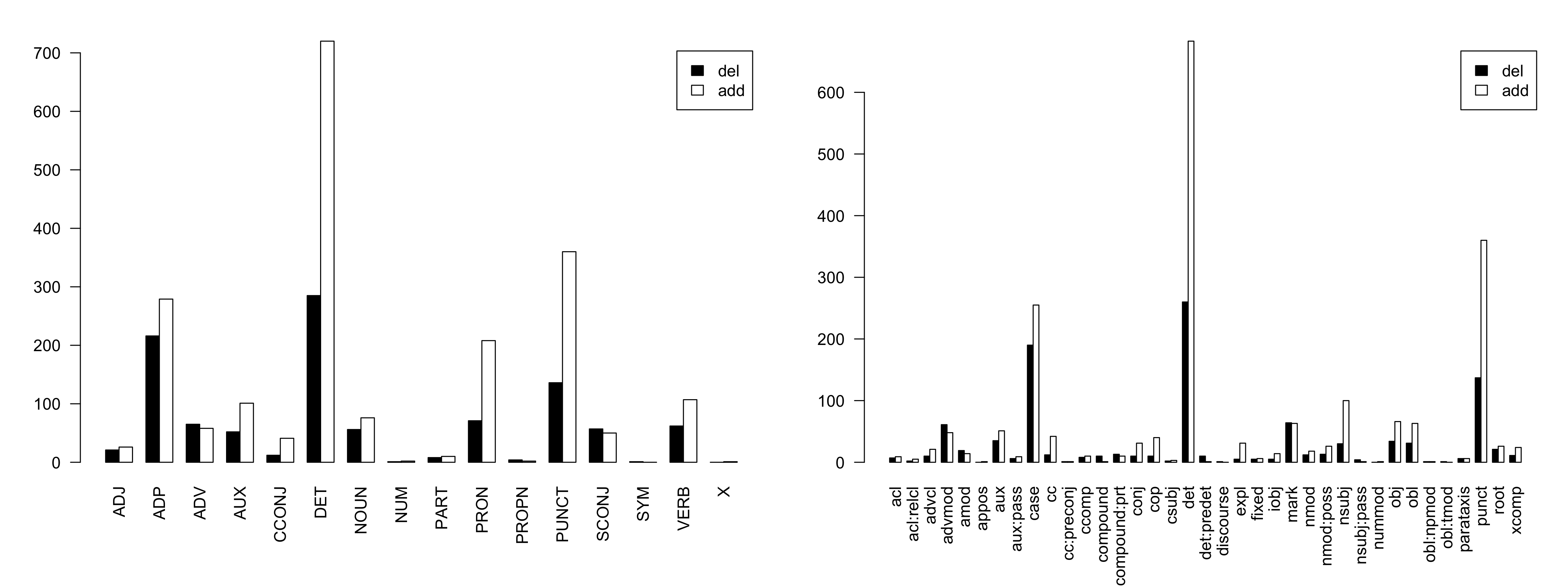}
		\caption{Left: POS tags of words deleted or added in corrected sentences in absolute counts; Right: edge labels of words deleted or added in corrected sentences in absolute counts (y). }
		
		\label{fig:add_del}
		\vspace{-0.5cm}
	\end{figure*}

	Figure \ref{fig:add_del} presents the most frequent addition and deletion types. Frequent POS tags are often frequently deleted or added POS tags, but not necessarily (e.g., nouns are almost twice as frequent as determiners). Additions are drastically more frequent than deletions for determiners, punctuation and pronouns and only slightly more for adpositions. Thus, we replicate the results that learners omit more than they add \citep{bryant-etal-2019-bea}, and give a detailed view on where they do not.
	
	It is often straightforward to connect major types of POS and edge-label additions and deletions: the relationship between DET and \texttt{det} is trivial, and missing/redundant adpositions mostly correspond to \texttt{mark} and \texttt{case}. Deletions and additions of lexical categories with more variegated syntactic functions (such as nouns and verbs) correspond to more varied edge labels. Generally, however, changes in edge labels and POS tags are found to be highly correlated both for additions and deletions (Cramer's V = \(0.78\) for both categories) and replacements (Cramer's V =  \(0.76\)). 
	
	Most prevalent types of replacements are presented in Table~\ref{table:pos_label_replacements}. These may suggest a direction to focus GEC efforts towards, a direction we explore in \S\ref{sec:uncommon}. 
	Full matrices are in appendix \S\ref{sec:matrices_TLE};
	incidentally, 44.4\% of the errors are POS SEs.

	\begin{table}[t]
		\begin{small}
			\centering
			\begin{tabular}{lccc|c}
				\toprule
				& A     & B     & C     & Native     \\ \midrule
				SCONJ & 0.804 & 0.864 & 0.923 & 0.942 \\
				DET   & 0.857 & 0.907 & 0.960 & 0.971 \\
				ADV   & 0.844 & 0.893 & 0.945 & 0.950 \\
				ADJ   & 0.875 & 0.923 & 0.962 & 0.972 \\
				ADP   & 0.891 & 0.935 & 0.969 & 0.976 \\
				PART  & 0.887 & 0.924 & 0.963 & 0.985 \\
				AUX   & 0.901 & 0.943 & 0.973 & 0.987 \\
				PROPN & 0.902 & 0.930 & 0.966 & 0.968 \\
				NUM   & 0.897 & 0.929 & 0.960 & 0.950 \\
				PRON  & 0.908 & 0.930 & 0.963 & 0.953 \\
				NOUN  & 0.934 & 0.963 & 0.983 & 0.983 \\
				CCONJ & 0.922 & 0.944 & 0.968 & 0.971 \\
				VERB  & 0.945 & 0.964 & 0.983 & 0.980 \\
				PUNCT & 0.978 & 0.980 & 0.990 & 0.981 \\
				\bottomrule
			\end{tabular}%
			
			\caption{Percentage of unchanged POS tags per type (rows) and proficiency level (columns) in the W\&I dataset. Proficiency levels are A-C where C is the most proficient, the last column is for native speakers. Sorted by the average of columns A-C.}
			\label{tab:levels_comp}
		\end{small}
		\vspace{-0.3cm}
	\end{table}
	
	Investigating SEs across levels (see Table \ref{tab:levels_comp}), we find the most error-prone SEs are among the least difficult to natives. However, on the easiest SEs advanced learners outperform natives. Being out of scope, we leave the details, as well as comparison between levels across datasets to Appendix \ref{ap:levels}. 
	
	\subsection{Russian} \label{sec:multi_language}
	
	To demonstrate the generality of the proposed approach, we apply \secl\ to RULEC, a corpus of learner Russian \citep{rozovskaya2019grammar}. 
	Russian syntax is characterised by pervasive agreement and complex rules of case selection for nouns. UD morphological features, parsed by UDPipe, make it possible to analyze learners' errors arising due to these phenomena; they are taken up in \S\ref{sec:morph}.
	
	\subsubsection{POS mismatches in learners' Russian}\label{ssec:russian-pos}
	
	\begin{figure}[t]
		\centering
		\includegraphics[width=0.5\textwidth]{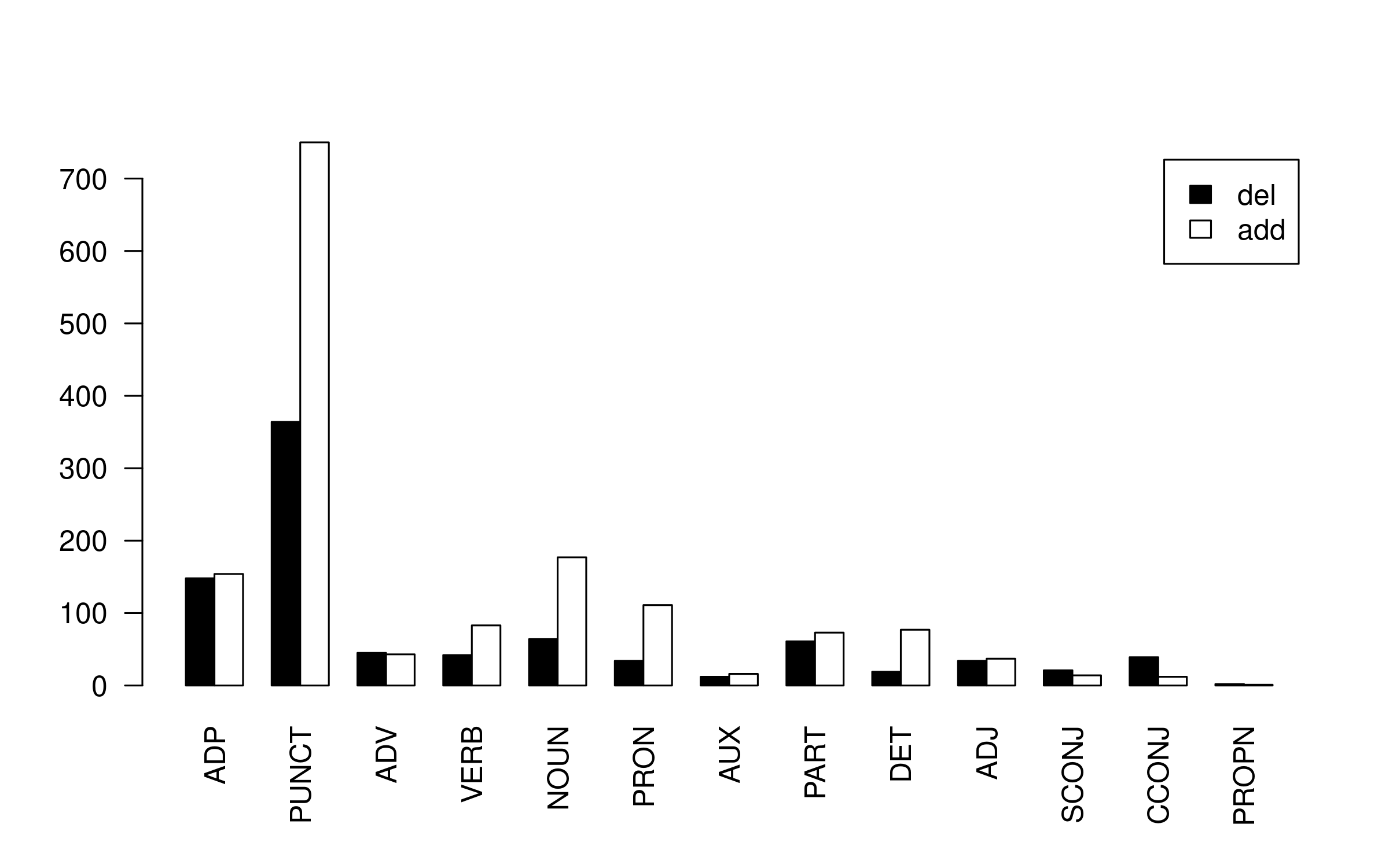}
		\caption{POS tags of words deleted or added in corrected Russian sentences.\label{fig:russian-pos}}
		\vspace{-0.2cm}
	\end{figure}
	
	An overview of POS additions and deletions is presented in Figure~\ref{fig:russian-pos}.
	Compared 
	to English (cf.\ \S\ref{sec:corpus_study_tle}), learners of Russian tend to more actively underuse nouns (177 additions vs.\ 64 deletions) and pronouns (111 additions vs.\ 34 deletions). 
	The latter may stem from Russian being a pro-drop language where subject pronouns can be omitted in certain contexts. The precise rules, however, are rather complicated, and it takes a lot of practice knowing when the result sounds felicitous \citep{zdorenko2010subject}.
	
	Most dominant types of POS replacements are similar (ADJ\ra NOUN, 80 cases; NOUN\ra ADJ, 75; VERB\ra NOUN, 65, PRON\ra DET, 50; NOUN\ra VERB, 45); 
	however, ADV\ra ADJ (66) and ADJ\ra ADV (51) are also prominent, which may be since adjectives and adverbs are more strictly distinct in Russian.
	
	\subsubsection{Morphological Features}\label{sec:morph}
	
	\begin{table}[t]
		\centering
		\begin{small}
			\begin{tabular}{lcccccc}
				\hline
				& \textbf{Acc} & \textbf{Dat} & \textbf{Gen} & \textbf{Ins} & \textbf{Loc} & \textbf{Nom} \\ 
				\textbf{Acc} & 0 & 46 & 132 & 43 & 96 & 40 \\
				\textbf{Dat} & 17 & 0 & 25 & 21 & 7 & 11 \\
				\textbf{Gen} & 78 & 45 & 0 & 71 & 73 & 83 \\
				\textbf{Ins} & 19 & 19 & 53 & 0 & 16 & 15 \\
				\textbf{Loc} & 66 & 14 & 62 & 12 & 0 & 7 \\
				\textbf{Nom} & 88 & 19 & 163 & 75 & 22 & 0 \\ \hline
			\end{tabular}
			\caption{Case corrections in nouns in Russian learners' sentences. Source(rows) against reference (columns).}\label{tab:case-rus}
		\end{small}
		\vspace{-0.5cm}
	\end{table}
	
	Russian possess a mildly
	complex conjugation and inflection system, which leaves a lot of room for errors even in cases when a correct POS is selected. The feature layer of UD makes it possible to identify these errors, which are dominated by three large classes: agreement errors (wrong person/number/gender features on verbs and wrong number/gender/case features on adjectives), case-assignment errors on nouns and pronouns, and verbal errors regarding aspect and voice.
	
	A breakdown of case-assignment errors for nouns is presented in Table~\ref{tab:case-rus}. It shows, among other things, that learners tend to use accusative and nominative cases in contexts where Russian demands the genitive case (which, in addition to the cross-linguistically frequent possessive meaning, also has numerous more subtle uses, e.g.\ in some types of negative sentences). Case-agreement errors on adjectives, on the other hand, tend to be more symmetric: 27 cases of an accusative case ending instead of a genitive vs.\ 19 cases of the converse error (see confusion matrices for case, gender, and number on adjectives in Appendix \S\ref{sec:matrices_morph}).
	
	The Russian verbal system also presents learners with several difficulties that our analysis echoes. Verbs fall into two aspectual classes (with \textit{perfective} verbs denoting completed actions and \textit{imperfective} verbs actions-in-progress and habitual actions), and it is difficult for learners with native languages lacking this distinction to use them correctly (e.g., the English phrase \textit{I went to work} will be translated differently depending on whether it is modified by \textit{yesterday} or \textit{every day}). The feature analysis shows that incorrect perfective verbs were changed into imperfective ones 210 times, and imperfective was changed into perfective 223 times.
	
	Another complication stems from the use of middle voice in Russian. English is dominated by \textit{labile} verbs denoting both spontaneous and caused actions (\textit{The cup broke} vs.\ \textit{I broke the cup}). In Russian, verbs for spontaneous actions are usually derived from transitive verbs (\textit{razbil}[\textsubscript{broke.smth}]\ra\textit{razbil\textbf{sja}}[\textsubscript{got.broken}]). Derived intransitive verbs are analyzed as ``middle voice'', and the inspection of verbal voice mismatches shows that the dominant type of error is the use of active voice instead of middle voice (108 cases vs.\ 45 cases of the converse error). 
	These findings can inform future efforts of GEC evaluation and development, in addressing these recurring error patterns.

	\section{Analyzing GEC System Outputs}\label{sec:system_outputs}

	To further showcase the utility of \secl, we demonstrate GEC system analysis with it.
	
	\subsection{Experiments with Leading GEC Systems}
	
	We use the outputs of several systems that participated in the 
	BEA2019 shared task \cite{bryant-etal-2019-bea}, namely: the winning system UEDIN-MS \citep{grundkiewicz2019neural}, as well as  KAKAO\&BRAIN \citep{choe2019neural}, SHUYAO \citep{xu2019erroneous}, CAMB-CUED \citep{stahlberg2019cueds}, and AIP-TOHOKU \citep{asano2019aip}, that were ranked second, fifth, eighth and ninth respectively.\footnote{The development outputs are published with the rest of our code, test outputs which we did not use are found in the contests' page. We thank Yoav Kantor for providing us with the development data.}
	We extract matrices for the system outputs using the same method as in \S\ref{sec:auto_parse}. Recall is bounded by the amount of predicted SEs, divided by their number in the gold standard. 
	The full matrices are given in Appendix  \S\ref{sec:out_matrices}.

	\begin{table}[tpb]
		\begin{small}
			\begin{tabular}{@{}lccc@{}}
				\toprule
				& AIP-TOHOKU & UEDIN-MS & GOLD \\
				\midrule
				ADJ\ra ADV & 18     & 21        & 55    \\
				ADJ\ra NOUN & 37    & 46        & 105   \\
				ADJ\ra VERB    & 21    & 33        & 59    \\
				NOUN \ra VERB & 76     & 87        & 142   \\
				VERB\ra ADJ & 17     & 21       & 38    \\
				NUM\ra DET  & 3      & 2        & 5     \\
				PART\ra DET & 0      & 3        & 1   \\
				\bottomrule
			\end{tabular}%
			\caption{AIP-TOHOKU, UEDIN-MS and Gold annotation changes (correct or not) on selected replacement types of SEs in absolute counts. The non-uniform behaviour of the systems over types indicates that \secl\ produces meaningful results.\label{tab:unique_outputs}}
		\end{small}
		\vspace{-0.5cm}
	\end{table}

	Our results in Table \ref{tab:unique_outputs} show that the top-ranking UEDIN-MS makes consistently more changes in general and per source SE than AIP-TOHOKU ranked 9th, but less than CAMB-CUED, ranked 8th (found in appendix \S\ref{sec:out_matrices}). 
	However, there is no correlation between the number of SE changes in general or per source tag and the rank of the system (both yield a partial-order Kendall $\tau=0$). This is in line with \citeauthor{choshen2018conservatism}'s (\citeyear{choshen2018conservatism}), who found no relation between a system's performance and its conservatism. 
	
	Some SE types are harder for the examined systems than others. There is a slight negative correlation ($r=-0.16$) between the average recall bound of the systems and frequency in the gold standard. For example, pronouns are well addressed across systems, with 65\% average 
	recall bound, while numerals are less so, with 43\%. This implies that less frequent corrections are handled less well, but also opens room for improvement. An example is numerals and coordinating conjunctions, which are handled less well.

	\begin{table}[htb]
		\centering
		\begin{small}
			\begin{tabular}{lccc}
				\toprule
				& UEDIN-MS & GOLD & Ratio (\%) \\
				\midrule
				CCONJ   & 71       & 158  & 44.9      \\
				NUM     & 22       & 48   & 45.8      \\
				SCONJ   & 114      & 233  & 48.9      \\
				AUX     & 153      & 310  & 49.4      \\
				VERB    & 202      & 405  & 49.9      \\
				ADP     & 232      & 461  & 50.3      \\
				PROPN   & 62       & 114  & 54.4      \\
				NOUN    & 346      & 618  & 56.0      \\
				ADV     & 273      & 472  & 57.8      \\
				PART    & 97       & 166  & 58.4      \\
				PUNCT   & 116      & 191  & 60.7      \\
				ADJ     & 283      & 462  & 61.3      \\
				DET     & 348      & 568  & 61.3      \\
				PRON    & 367      & 584  & 62.8      \\
				\midrule
				Overall & 2686     & 4790 & 56.1     \\
			\end{tabular}
			\caption{Amount of syntactic changes per source POS tags for UEDIN-MS and the gold standard in absolute counts. The ratio is an upper bound on the recall.}
			\label{tab:UEDIN_vs_gold}
		\end{small}
		\vspace{-0.5cm}
	\end{table}

	We further revisit the replacement types discussed in \S\ref{sec:corpus_study_tle} and compute the recall upper bound for the top system UEDIN-MS (Table \ref{tab:unique_outputs} and aggregation per source POS in \ref{tab:UEDIN_vs_gold}). Putting aside the rare types NUM\ra DET and PART\ra DET, we find that UEDIN-MS tackles considerably more SEs than Grammarly (see \S\ref{sec:uncommon} below and Table~\ref{tab:Grammarly_performance}). 
	However, the recall bound is not uniform across types, where two types receive very low recall (ADJ\ra ADV with 38\% and ADJ\ra NOUN with 44\%), indicating these as potential directions for future work. 
	
	The bound over all SEs for UEDIN-MS is
	56\% (50\% on the subset of errors discussed above). If we assume the precision on SE is similar to the overall reported precision (72\%), we may conclude that recall for SEs is around 40\%. For comparison, the overall reported recall is 60\%, which suggests that SEs are harder on average than non-SEs, underscoring the value in classifying them.

	\subsection{Prospects for Improving GEC using Fine-grained SE Classification}\label{sec:uncommon}
	
	To demonstrate the benefits of fine-grained SE categories, 
	we analyze several SE types involving word replacements that are not as prevalent in TLE as additions and deletions of determiners and prepositions, but are still recurrent and form a closed-class that is likely to be addressable through designated GEC modules. We also consider the open-class VERB\ra ADJ for comparison.
	We examine the performance of a leading GEC tool, \href{https://app.grammarly.com/}{Grammarly}
	in handling these types and analyze the capabilities of end-to-end systems in \S\ref{sec:system_outputs}. 
	
	
	\paragraph{NUM\ra DET.} Almost all examples include \textit{One} instead of \textit{any} or \textit{another}. Example: \textit{Technology is also important in \sout{one} another area for me.}
	
	\paragraph{PART\ra DET.} All examples show \textit{not} used instead of \textit{no}. Example: \textit{There was \sout{not} no discount}.
	
	\paragraph{ADJ\ra NOUN.} Such replacements mostly involve quantifiers (\textit{many} \ra\textit{a lot of}, \textit{small} \ra\textit{a few}) which constitute a closed class. Example: \textit{But some schools in my country don't allow \sout{many} a lot of things.} 
	Another subtype of this SE is more open-ended in that it involves using adjectives instead of morphologically related nouns (e.g., \textit{joyful}\ra\textit{joy}, \textit{late} \ra\textit{lateness}). Example: \textit{Second, there should be friends and family members in the home to provide \sout{joyful} joy and fun.} These SEs should be easy to detect because derivational relations are mostly transparent. However, there are a handful of harder cases (e.g., \textit{I felt like a \sout{dumb} fool}) whose correction demands more nuanced lexical knowledge.
	
	\paragraph{NOUN\ra VERB.} This SE type usually involves a morphologically-related form  (\textit{entrance}\ra\textit{enter}, \textit{product}\ra\textit{produce}). However, some of the examples of this type are ambiguous due to English zero-derivation of deverbal nouns.
	Cf. \textit{I love sleep in tents}, where \textit{to sleep} and \textit{sleeping} are both valid corrections found in the corpus. Example: \textit{When we \sout{entrance} entered the place our problems began.}
	
	\paragraph{VERB\ra ADJ.}
	Those replacements are diverse and often include large changes, mostly when the original sentence uses a completely wrong form of expression. Errors involving a verb rather a passive participle, which acts as an adjective, are also frequent (\textit{trust was broken}\ra\textit{betrayed trust}, \textit{have conscience}\ra\textit{are aware}, \textit{problems involve with} \ra\textit{problems involved with}).

	\begin{table*}[tpb]
		\centering
		\begin{small}
			\begin{tabular}{@{}clrrrrr@{}}
				\toprule
				\multicolumn{1}{l}{} &  & \multicolumn{1}{l}{Amount} & \multicolumn{1}{l}{Detected} & \multicolumn{1}{l}{Valid} & \multicolumn{1}{l}{Precision} & \multicolumn{1}{l}{Recall} \\ \midrule
				\multirow{4}{*}{Add} & None\ra ADV & 58 & 0 & 0 & 0\% & 0\% \\
				& None\ra DET & 44 & 13 & 10 & 77\% & 23\% \\
				& None\ra PRON & 120 & 4 & 1 & 25\% & 1\% \\
				& None\ra VERB & 107 & 0 & 0 & 0\% & 0\% \\
				\multirow{4}{*}{Delete} & ADV\ra None & 64 & 3 & 0 & 0\% & 0\% \\
				& DET\ra None & 49 & 20 & 20 & 100\% & 41\% \\
				& Pron\ra None & 71 & 4 & 4 & 100\% & 6\% \\
				& VERB\ra None & 41 & 4 & 0 & 0\% & 0\% \\
				\multirow{7}{*}{Replace} & ADJ\ra ADV & 101 & 19 & 12 & 63\% & 12\% \\
				& ADJ\ra NOUN & 45 & 4 & 2 & 50\% & 4\% \\
				& ADJ\ra VERB & 18 & 0 & 0 & 0\% & 0\% \\
				& NOUN\ra VERB & 44 & 10 & 8 & 80\% & 18\% \\
				& VERB\ra ADJ & 26 & 3 & 0 & 0\% & 0\% \\
				& NUM\ra DET & 7 & 0 & 0 & 0\% & 0\% \\
				& PART\ra DET & 6 & 4 & 4 & 100\% & 67\% \\ \cmidrule(l){2-7} 
			\end{tabular}%
			\caption{Grammarly's performance on selected SE types in absolute counts. The varying behaviour per type indicates the separation to types is meaningful.\label{tab:Grammarly_performance}
				\vspace{-.5cm}}
		\end{small}
	\end{table*}

	\paragraph{}
	We turn to analyzing Grammarly's performance on the types discussed,
	as well as the four most frequent SE types of deletions, additions, and replacements (two of which are not among the above types). 
	Grammarly's performance is of particular interest due to its reliance on designated modules (classifiers and rules) for addressing specific error types. Our results thus demonstrate how such a system may benefit from uncovering error types that can be addressed by integrating additional modules.
	
	We manually annotate whether the edit in question is at all detected and whether it is validly corrected by Grammarly. As Grammarly may offer more than a single correction, we deem correct any case where at least one of the corrections is valid.
	
	Results (Table \ref{tab:Grammarly_performance}) indicate that Grammarly fares poorly in addressing SEs, with the possible exception of superfluous determiners. Indeed, in many of the cases, only a small portion of the SEs was detected. While it is possible that Grammarly tends to overlook such cases because of the dominance of punctuation, spelling, and determiner errors in learner language, some of the types here involve only a handful of lexemes, suggesting that targeted treatment or data augmentation may be effective.

	\vspace{-.1cm}
	\section{Related Work}\label{sec:related}
	\vspace{-.1cm}
	
	Error types are often used to improve performance and evaluation in GEC. Taxonomies have been used to construct classifiers and rule-based engines to correct specific error types \cite[e.g.,][]{rozovskaya2014correcting,farra2014generalized,zheng2018you}. 
	When using end-to-end systems, balancing the distribution of errors in the train and test sets has been shown to improve results \cite{junczys2018approaching}. 
	Ensembling black-box systems relying on per-type performance has been shown superior to each system's performance and over average ensembling \citep{kantor2019learning}.
	Augmenting the training data with synthetic errors of a particular type is effective for improving performance on that type \cite{belinkov2018synthetic}. 
	The classification of grammatical error types is also used to analyze system performance \citep[e.g.,][]{Lichtarge2019CorporaGF}.  
	\citet{choshen2018maege,choshen2018conservatism} showed that current systems and evaluation measures essentially ignore some error types, suggesting that targeted evaluation of these types may be needed. 
	
	To date, several error taxonomies have been proposed and applied for annotating errors in major English learner-language corpora \citep[][{\it inter alia}]{bryant-etal-2019-bea, dahlmeier2013building,nicholls2003cambridge}. There has been interest lately in other languages, for which different datasets and taxonomies were created \citep{rozovskaya2019grammar, rao2018overview, zaghouani2014large}.
	However, different taxonomies are used by different corpora, based on commonly observed error types in the target domain and language, which impedes direct comparison across corpora. Moreover, these taxonomies are not formulated based on a specific theory or annotation scheme for morphosyntactic representation, which may promote accessibility to non-experts but often leads to non-uniform terminology and difficulty in leveraging available NLP tools.
	
	Another automatic type classification was suggested apart from ERRANT. \citet{swanson2012correction} trained a log-linear model to predict types defined by \citet{nicholls2003cambridge}. This taxonomy resembles ours in that it uses grammatical categories (POS tags), but differs in that it only distinguishes types based on the POS tag of the correction and not of the source sentence. Moreover, relying solely on POS tags yields difficulties in classifying constructions that involve more than a single word. For such cases, it defines specialized error types, such as {\it Incorrect Argument Structure}, which serves as a residual category for argument structure errors that cannot be accounted for by adposition or agreement errors. However, unlike \secl,  it does not provide any information as to what particular incorrect argument structure was used or how it should be corrected.
	
	\citet{choshen2018usim} used a semantic annotation\cite{Abend2013UniversalCC} to show semantics, unlike syntax is kept upon changes.
	UD was previously used in GEC in the TLE corpus and in a learner language parser \citep[e.g.,][]{sakaguchi2017error} (we do not apply their parser, as it is made specifically for English, and might alter the origin parse).

	\vspace{-0.1cm}
	\section{Conclusion}\label{sec:conclusion}
	\vspace{-0.1cm}
	We presented \secl, a novel method for classifying SEs based on UD parses of learner text and its correction. We show that \secl\ provides a detailed picture of the prevalence of different SEs in two languages, and can be straightforwardly automated. We further show that the method manages to classify about 60\% of the unclassified edits by ERRANT, the standard tool for error classification, and provides useful complementary information for many of the classified edits. 
	
	Future work will combine \secl\ and ERRANT into a single tool for English error classification (work in this direction has already begun). The experiments we presented show that several leading GEC systems of different types make errors of types that are not well-addressed by current systems. These results can inform the future development of tailored solutions for these cases. 
	
	\section*{Acknowledgments}
	We thank Yarden Gavish for her help with coding and data preparation assignments. 
	This work was supported by the Israel Science Foundation (grant no. 929/17).
	Leshem Choshen is supported by the Clore Foundation.
	
	\bibliography{emnlp-ijcnlp-2019}

\begin{thebibliography}{34}
\expandafter\ifx\csname natexlab\endcsname\relax\def\natexlab#1{#1}\fi

\bibitem[{Abend and Rappoport(2013)}]{Abend2013UniversalCC}
Omri Abend and A.~Rappoport. 2013.
\newblock Universal conceptual cognitive annotation ({UCCA}).
\newblock In \emph{ACL}.

\bibitem[{Asano et~al.(2019)Asano, Mita, Mizumoto, and Suzuki}]{asano2019aip}
Hiroki Asano, Masato Mita, Tomoya Mizumoto, and Jun Suzuki. 2019.
\newblock \href {https://doi.org/10.18653/v1/W19-4418} {The {AIP}-tohoku system
  at the {BEA}-2019 shared task}.
\newblock In \emph{Proceedings of the Fourteenth Workshop on Innovative Use of
  NLP for Building Educational Applications}, pages 176--182, Florence, Italy.
  Association for Computational Linguistics.

\bibitem[{Belinkov and Bisk(2018)}]{belinkov2018synthetic}
Yonatan Belinkov and Yonatan Bisk. 2018.
\newblock Synthetic and natural noise both break neural machine translation.
\newblock \emph{ICLR}.

\bibitem[{Berzak et~al.(2016)Berzak, Kenney, Spadine, Wang, Lam, Mori, Garza,
  and Katz}]{berzak2016universal}
Yevgeni Berzak, Jessica Kenney, Carolyn Spadine, Jing~Xian Wang, Lucia Lam,
  Keiko~Sophie Mori, Sebastian Garza, and Boris Katz. 2016.
\newblock Universal {D}ependencies for learner english.
\newblock In \emph{Proceedings of the 54th Annual Meeting of the Association
  for Computational Linguistics (Volume 1: Long Papers)}, volume~1, pages
  737--746.

\bibitem[{Boyd(2018)}]{boyd2018using}
Adriane Boyd. 2018.
\newblock \href {https://doi.org/10.18653/v1/W18-6111} {Using {W}ikipedia edits
  in low resource grammatical error correction}.
\newblock In \emph{Proceedings of the 2018 {EMNLP} Workshop W-{NUT}: The 4th
  Workshop on Noisy User-generated Text}, pages 79--84, Brussels, Belgium.
  Association for Computational Linguistics.

\bibitem[{Bryant et~al.(2019)Bryant, Felice, Andersen, and
  Briscoe}]{bryant-etal-2019-bea}
Christopher Bryant, Mariano Felice, {\O}istein~E. Andersen, and Ted Briscoe.
  2019.
\newblock \href {https://doi.org/10.18653/v1/W19-4406} {The {BEA}-2019 shared
  task on grammatical error correction}.
\newblock In \emph{Proceedings of the Fourteenth Workshop on Innovative Use of
  NLP for Building Educational Applications}, pages 52--75, Florence, Italy.
  Association for Computational Linguistics.

\bibitem[{Bryant et~al.(2017)Bryant, Felice, and Briscoe}]{bryant2017automatic}
Christopher Bryant, Mariano Felice, and Ted Briscoe. 2017.
\newblock Automatic annotation and evaluation of error types for grammatical
  error correction.
\newblock In \emph{ACL}.

\bibitem[{Choe et~al.(2019)Choe, Ham, Park, and Yoon}]{choe2019neural}
Yo~Joong Choe, Jiyeon Ham, Kyubyong Park, and Yeoil Yoon. 2019.
\newblock \href {https://doi.org/10.18653/v1/W19-4423} {A neural grammatical
  error correction system built on better pre-training and sequential transfer
  learning}.
\newblock In \emph{Proceedings of the Fourteenth Workshop on Innovative Use of
  NLP for Building Educational Applications}, pages 213--227, Florence, Italy.
  Association for Computational Linguistics.

\bibitem[{Choshen and Abend(2018{\natexlab{a}})}]{choshen2018maege}
Leshem Choshen and Omri Abend. 2018{\natexlab{a}}.
\newblock Automatic metric validation for grammatical error correction.
\newblock In \emph{Proceedings of the 56th Annual Meeting of the Association
  for Computational Linguistics (Volume 1: Long Papers)}.

\bibitem[{Choshen and Abend(2018{\natexlab{b}})}]{choshen2018conservatism}
Leshem Choshen and Omri Abend. 2018{\natexlab{b}}.
\newblock Inherent biases in reference-based evaluation for grammatical error
  correction and text simplification.
\newblock In \emph{Proceedings of the 56th Annual Meeting of the Association
  for Computational Linguistics (Volume 1: Long Papers)}.

\bibitem[{Choshen and Abend(2018{\natexlab{c}})}]{choshen2018usim}
Leshem Choshen and Omri Abend. 2018{\natexlab{c}}.
\newblock \href {http://aclweb.org/anthology/N18-2020} {Reference-less measure
  of faithfulness for grammatical error correction}.
\newblock In \emph{Proceedings of the 2018 Conference of the North American
  Chapter of the Association for Computational Linguistics: Human Language
  Technologies}.

\bibitem[{Dahlmeier et~al.(2013)Dahlmeier, Ng, and Wu}]{dahlmeier2013building}
Daniel Dahlmeier, Hwee~Tou Ng, and Siew~Mei Wu. 2013.
\newblock Building a large annotated corpus of learner english: The nus corpus
  of learner english.
\newblock In \emph{Proceedings of the eighth workshop on innovative use of NLP
  for building educational applications}, pages 22--31.

\bibitem[{Dorr(1994)}]{dorr1994machine}
Bonnie~J. Dorr. 1994.
\newblock Machine translation divergences: a formal description and proposed
  solution.
\newblock \emph{Computational linguistics}, 20(4):597--635.

\bibitem[{Farra et~al.(2014)Farra, Tomeh, Rozovskaya, and
  Habash}]{farra2014generalized}
Noura Farra, Nadi Tomeh, Alla Rozovskaya, and Nizar Habash. 2014.
\newblock Generalized character-level spelling error correction.
\newblock In \emph{Proceedings of the 52nd Annual Meeting of the Association
  for Computational Linguistics (Volume 2: Short Papers)}, volume~2, pages
  161--167.

\bibitem[{Grundkiewicz et~al.(2019)Grundkiewicz, Junczys-Dowmunt, and
  Heafield}]{grundkiewicz2019neural}
Roman Grundkiewicz, Marcin Junczys-Dowmunt, and Kenneth Heafield. 2019.
\newblock \href {https://doi.org/10.18653/v1/W19-4427} {Neural grammatical
  error correction systems with unsupervised pre-training on synthetic data}.
\newblock In \emph{Proceedings of the Fourteenth Workshop on Innovative Use of
  NLP for Building Educational Applications}, pages 252--263, Florence, Italy.
  Association for Computational Linguistics.

\bibitem[{Junczys-Dowmunt et~al.(2018)Junczys-Dowmunt, Grundkiewicz, Guha, and
  Heafield}]{junczys2018approaching}
Marcin Junczys-Dowmunt, Roman Grundkiewicz, Shubha Guha, and Kenneth Heafield.
  2018.
\newblock Approaching neural grammatical error correction as a low-resource
  machine translation task.
\newblock In \emph{NAACL-HLT}.

\bibitem[{Kantor et~al.(2019)Kantor, Katz, Choshen, Naftali, and
  Slonim}]{kantor2019learning}
Yoav Kantor, Yoav Katz, Leshem Choshen, Naftali Naftali, and Noam Slonim. 2019.
\newblock Learning to combine grammatical error corrections.
\newblock In \emph{BEA 2019 Shared Task: Grammatical Error Correction}.

\bibitem[{Lichtarge et~al.(2019)Lichtarge, Alberti, Kumar, Shazeer, Parmar, and
  Tong}]{Lichtarge2019CorporaGF}
Jared Lichtarge, Christopher Alberti, Shankar Kumar, Noam Shazeer, Niki Parmar,
  and Simon Tong. 2019.
\newblock Corpora generation for grammatical error correction.
\newblock In \emph{NAACL-HLT}.

\bibitem[{Mizumoto et~al.(2012)Mizumoto, Hayashibe, Komachi, Nagata, and
  Matsumoto}]{Mizumoto2012TheEO}
Tomoya Mizumoto, Yuta Hayashibe, Mamoru Komachi, Masaaki Nagata, and Yuji
  Matsumoto. 2012.
\newblock The effect of learner corpus size in grammatical error correction of
  esl writings.
\newblock In \emph{COLING}.

\bibitem[{Nicholls(2003)}]{nicholls2003cambridge}
Diane Nicholls. 2003.
\newblock The cambridge learner corpus: Error coding and analysis for
  lexicography and elt.
\newblock In \emph{Proceedings of the Corpus Linguistics 2003 conference},
  volume~16, pages 572--581.

\bibitem[{Nikolaev et~al.(2020)Nikolaev, Arviv, Karidi, Kenneth, Mitnik,
  Saeboe, and Abend}]{nikolaev2020fine}
Dmitry Nikolaev, Ofir Arviv, Taelin Karidi, Neta Kenneth, Veronika Mitnik,
  Lilja~Maria Saeboe, and Omri Abend. 2020.
\newblock \href {https://www.aclweb.org/anthology/2020.acl-main.109}
  {Fine-grained analysis of cross-linguistic syntactic divergences}.
\newblock In \emph{Proceedings of the 58th Annual Meeting of the Association
  for Computational Linguistics}, pages 1159--1176, Online. Association for
  Computational Linguistics.

\bibitem[{Nivre et~al.(2016)Nivre, de~Marneffe, Ginter, Goldberg, Hajic,
  Manning, McDonald, Petrov, Pyysalo, Silveira, Tsarfaty, and
  Zeman}]{nivre2016universal}
Joakim Nivre, Marie-Catherine de~Marneffe, Filip Ginter, Yoav Goldberg, Jan
  Hajic, Christopher~D. Manning, Ryan McDonald, Slav Petrov, Sampo Pyysalo,
  Natalia Silveira, Reut Tsarfaty, and Daniel Zeman. 2016.
\newblock \href {https://nlp.stanford.edu/pubs/nivre2016ud.pdf} {Universal
  {D}ependencies v1: A multilingual treebank collection}.
\newblock In \emph{Proc. of LREC}.

\bibitem[{Rao et~al.(2018)Rao, Gong, Zhang, and Xun}]{rao2018overview}
Gaoqi Rao, Qi~Gong, Baolin Zhang, and Endong Xun. 2018.
\newblock Overview of nlptea-2018 share task chinese grammatical error
  diagnosis.
\newblock In \emph{Proceedings of the 5th Workshop on Natural Language
  Processing Techniques for Educational Applications}, pages 42--51.

\bibitem[{Rozovskaya and Roth(2019)}]{rozovskaya2019grammar}
Alla Rozovskaya and Dan Roth. 2019.
\newblock Grammar error correction in morphologically rich languages: The case
  of russian.
\newblock \emph{Transactions of the Association for Computational Linguistics},
  7:1--17.

\bibitem[{Rozovskaya et~al.(2014)Rozovskaya, Roth, and
  Srikumar}]{rozovskaya2014correcting}
Alla Rozovskaya, Dan Roth, and Vivek Srikumar. 2014.
\newblock Correcting grammatical verb errors.
\newblock In \emph{Proceedings of the 14th Conference of the European Chapter
  of the Association for Computational Linguistics}, pages 358--367.

\bibitem[{Sakaguchi et~al.(2017)Sakaguchi, Post, and
  Van~Durme}]{sakaguchi2017error}
Keisuke Sakaguchi, Matt Post, and Benjamin Van~Durme. 2017.
\newblock Error-repair dependency parsing for ungrammatical texts.
\newblock In \emph{Proceedings of the 55th Annual Meeting of the Association
  for Computational Linguistics (Volume 2: Short Papers)}, pages 189--195.

\bibitem[{Stahlberg and Byrne(2019)}]{stahlberg2019cueds}
Felix Stahlberg and Bill Byrne. 2019.
\newblock \href {https://doi.org/10.18653/v1/W19-4417} {The {CUED}{'}s
  grammatical error correction systems for {BEA}-2019}.
\newblock In \emph{Proceedings of the Fourteenth Workshop on Innovative Use of
  NLP for Building Educational Applications}, pages 168--175, Florence, Italy.
  Association for Computational Linguistics.

\bibitem[{Straka et~al.(2016)Straka, Hajic, and
  Strakov{\'a}}]{Straka2016UDPipeTP}
Milan Straka, Jan Hajic, and Jana Strakov{\'a}. 2016.
\newblock Udpipe: Trainable pipeline for processing conll-u files performing
  tokenization, morphological analysis, pos tagging and parsing.
\newblock In \emph{LREC}.

\bibitem[{Swanson and Yamangil(2012)}]{swanson2012correction}
Ben Swanson and Elif Yamangil. 2012.
\newblock \href {https://www.aclweb.org/anthology/N12-1037} {Correction
  detection and error type selection as an {ESL} educational aid}.
\newblock In \emph{Proceedings of the 2012 Conference of the North {A}merican
  Chapter of the Association for Computational Linguistics: Human Language
  Technologies}, pages 357--361, Montr{\'e}al, Canada. Association for
  Computational Linguistics.

\bibitem[{Xu et~al.(2019)Xu, Zhang, Chen, and Qin}]{xu2019erroneous}
Shuyao Xu, Jiehao Zhang, Jin Chen, and Long Qin. 2019.
\newblock \href {https://doi.org/10.18653/v1/W19-4415} {Erroneous data
  generation for grammatical error correction}.
\newblock In \emph{Proceedings of the Fourteenth Workshop on Innovative Use of
  NLP for Building Educational Applications}, pages 149--158, Florence, Italy.
  Association for Computational Linguistics.

\bibitem[{Zaghouani et~al.(2014)Zaghouani, Mohit, Habash, Obeid, Tomeh,
  Rozovskaya, Farra, Alkuhlani, and Oflazer}]{zaghouani2014large}
Wajdi Zaghouani, Behrang Mohit, Nizar Habash, Ossama Obeid, Nadi Tomeh, Alla
  Rozovskaya, Noura Farra, Sarah Alkuhlani, and Kemal Oflazer. 2014.
\newblock \href
  {http://www.lrec-conf.org/proceedings/lrec2014/pdf/956_Paper.pdf} {Large
  scale {A}rabic error annotation: Guidelines and framework}.
\newblock In \emph{Proceedings of the Ninth International Conference on
  Language Resources and Evaluation ({LREC}'14)}, Reykjavik, Iceland. European
  Language Resources Association (ELRA).

\bibitem[{Zdorenko(2010)}]{zdorenko2010subject}
Tatiana Zdorenko. 2010.
\newblock Subject omission in russian: a study of the russian national corpus.
\newblock In Stefan~Th. Gries, Stefanie Wulff, and Mark Davies, editors,
  \emph{Corpus-linguistic applications. Current studies, new directions}, pages
  119--133. Brill Rodopi.

\bibitem[{Zeman et~al.(2018)Zeman, Haji{\v{c}}, Popel, Potthast, Straka,
  Ginter, Nivre, and Petrov}]{zeman2018conll}
Daniel Zeman, Jan Haji{\v{c}}, Martin Popel, Martin Potthast, Milan Straka,
  Filip Ginter, Joakim Nivre, and Slav Petrov. 2018.
\newblock Conll 2018 shared task: multilingual parsing from raw text to
  universal dependencies.
\newblock In \emph{Proceedings of the CoNLL 2018 Shared Task: Multilingual
  Parsing from Raw Text to Universal Dependencies}, pages 1--21.

\bibitem[{Zheng et~al.(2018)Zheng, Napoles, Tetreault, and
  Omelianchuk}]{zheng2018you}
Junchao Zheng, Courtney Napoles, Joel Tetreault, and Kostiantyn Omelianchuk.
  2018.
\newblock How do you correct run-on sentences it’s not as easy as it seems.
\newblock In \emph{Proceedings of the 2018 EMNLP Workshop W-NUT: The 4th
  Workshop on Noisy User-generated Text}, pages 33--38.

\end{thebibliography}


\begin{thebibliography}{1}
\expandafter\ifx\csname natexlab\endcsname\relax\def\natexlab#1{#1}\fi

\bibitem[{Little(2006)}]{Little2006TheCE}
David Little. 2006.
\newblock The common european framework of reference for languages: Content,
  purpose, origin, reception and impact.
\newblock In \emph{Cambridge Journal}.

\end{thebibliography}
	\bibliographystyle{acl_natbib}
	
	\appendix

	\section{Classifying SEs across Learner Levels} \label{ap:levels}
	
	The W\&I dataset includes both corrections of texts by native speakers and by different levels of non-native speakers, we use that to analyze changes across proficiency levels. We extract the distributions of SE types from the train set for each proficiency level A--C, where A is the lowest, and for the native speakers' text. 
	We compare across levels what percentage of words of a given POS tag are SEs (i.e., changed to a different POS in the correction). Results are presented in Table~\ref{ap_tab:levels_comp}; full confusion matrices are given in Appendix \S\ref{sec:wi_matrices}.
	
	As one might expect, the more proficient the learners, the fewer SEs they make. This holds for learners across all POS types. 
	Native speakers generally make fewer errors than advanced learners, but the trend is mixed. Surprisingly, native speakers rarely make SEs in POS tags that are the most error-prone for learners (top of Table \ref{ap_tab:levels_comp}). Conversely, native speakers generally do not do better than advanced learners in the least error-prone POS tags (bottom of the table). It is important to note that as the non natives have less data, the borderline changes positive and negative changes might be considered comparable rather than significant changes to one way or another. Further study of this phenomenon is deferred to future work.

	
	
	
	Finally, we compare the syntactic proficiency of the learner language in TLE and the A--C proficiency levels of W\&I. We compute the percentage of POS tags of each type that were altered in the correction. We find the syntactic proficiency of TLE to be not too high, in most cases between level A and B. Those results agree with the prior knowledge that TLE texts are taken from B1-B2 CEFR 
	levels, despite the difference in the native tongues of the learners, which adds noise to the comparison.
	This result thus provides further validation for \secl. 

	\begin{table}[t]
		\begin{small}
			\centering
			\begin{tabular}{lccc|c}
				\toprule
				& A     & B     & C     & Native     \\ \midrule
				SCONJ & 0.804 & 0.864 & 0.923 & 0.942 \\
				DET   & 0.857 & 0.907 & 0.960 & 0.971 \\
				ADV   & 0.844 & 0.893 & 0.945 & 0.950 \\
				ADJ   & 0.875 & 0.923 & 0.962 & 0.972 \\
				ADP   & 0.891 & 0.935 & 0.969 & 0.976 \\
				PART  & 0.887 & 0.924 & 0.963 & 0.985 \\
				AUX   & 0.901 & 0.943 & 0.973 & 0.987 \\
				PROPN & 0.902 & 0.930 & 0.966 & 0.968 \\
				NUM   & 0.897 & 0.929 & 0.960 & 0.950 \\
				PRON  & 0.908 & 0.930 & 0.963 & 0.953 \\
				NOUN  & 0.934 & 0.963 & 0.983 & 0.983 \\
				CCONJ & 0.922 & 0.944 & 0.968 & 0.971 \\
				VERB  & 0.945 & 0.964 & 0.983 & 0.980 \\
				PUNCT & 0.978 & 0.980 & 0.990 & 0.981 \\
				\bottomrule
			\end{tabular}%
			
			\caption{Percentage of unchanged POS tags per type (rows) and proficiency level (columns) in the W\&I dataset. Proficiency levels are A-C where C is the most proficient, the last column is for native speakers. Sorted by the average of columns A-C.}
			\label{ap_tab:levels_comp}
		\end{small}
		\vspace{-0.3cm}
	\end{table}
	
	\FloatBarrier
	\section{Examples of Prevalent Types of SEs Involving Replacements}\label{ssec:replacement_examples}
	
	\subsection{POS Changes\footnote{When sentences contained other errors the corrected tokens were used for clarity of presentation.}}
	\FloatBarrier
	
	\paragraph{NOUN~\ra~VERB}
	\begin{itemize}
		\item \textit{When we \sout{entrance} entered the place our problems began}.
		\item \textit{I mean it\sout{'s conflict} conflicts with your plan, what a pity!}
	\end{itemize}
	
	\paragraph{NOUN~\ra~ADJ}
	
	\begin{itemize}
		\item \textit{These things are very \sout{convenience} convenient; on the other hand, there are lots of disadvantage points}.
		\item \textit{In this way the \sout{environment} environmental pollution will increase more and more.}
	\end{itemize}
	
	\paragraph{ADJ~\ra~NOUN}
	
	\begin{itemize}
		\item \textit{So while I am at the Camp I like to take \sout{many} a lot of photographs and climb a mountain}.
		\item \textit{He has known research in the \sout{deeper} depths of his mind.}
	\end{itemize}
	
	\paragraph{VERB~\ra~NOUN}
	
	\begin{itemize}
		\item \textit{I think that I am a very fast \sout{swimming} swimmer but I don't have good style and technique.}
		\item \textit{Obviously, there is no point in saying ``Famous people have a right to their own \sout{live} lives.''}
	\end{itemize}
	
	\paragraph{VERB~\ra~ADJ}
	
	\begin{itemize}
		\item \textit{There was only a small sign \sout{stamped} stuck on the door, saying that it was ``closed for repairs''.}
		\item \textit{Sometimes you \sout{surprise} are surprised when you check the balance of your bank account.}
	\end{itemize}
	
	\paragraph{DET~\ra~PRON}
	
	\begin{itemize}
		\item \textit{I have a camera, \sout{which} it is called Minolta X300.}
		\item \textit{To finish, I wonder if I need to bring some money or if \sout{all} everything has been already paid for.}
	\end{itemize}
	
	\paragraph{PRON~\ra~DET}
	
	\begin{itemize}
		\item \textit{I was pleased to receive \sout{you} your letter recently.}
		\item \textit{I can say that the greatest ever invention is the invention of computers, \sout{that} which has affected both individuals and a society as a whole.}
	\end{itemize}
	
	\subsubsection{Edge-type Changes\footnote{In cases of clause-linkage errors, words participating in the relations are underlined; words inducing the relations are in boldface.}}
	
	\paragraph{compound~\ra~amod}
	
	\begin{itemize}
		\item \textit{Shopping is relaxing, above all on Saturdays when you have finished a \sout{work} working week and you are expecting a wonderful Sunday.}
		\item \textit{Secondly, turning to the end-of-conference party, it is planned to be in Wimbledon Common -- nice piece of \sout{England} English nature.}
	\end{itemize}
	
	\paragraph{cop~\ra~aux}
	
	\begin{itemize}
		\item \textit{From my point of view I think he was such a very bad actor that it \sout{was} would be impossible to fine another with the same characteristics.}
		\item \textit{It should \sout{be} have been a perfect evening out, but it was the opposite.}
	\end{itemize}
	
	\paragraph{xcomp~\ra~ccomp}
	
	\begin{itemize}
		\item \textit{It \textbf{made} people only \underline{spend} pocket-money} \ra \textit{It \textbf{meant} people only \underline{spent} pocket-money}.
		\item \textit{I went in London for a week of ``pure holiday'' thinking
			\textbf{to} \underline{have} some fun coming to see your musical show} \ra \textit{I went to London for a week of ``pure holiday'', thinking I would \underline{have} some fun going to see your musical show.}
	\end{itemize}
	
	\paragraph{obl~\ra~obj}
	
	\begin{itemize}
		\item \textit{I'm not so good in tennis but I would like to play it \textbf{for} \underline{practice}} \ra \textit{I'm not so good at tennis but I would like to play it to \textbf{get} some \underline{practice}.}
		\item \textit{However, we have to consider \sout{\textbf{about}} our human \underline{nature}.}
	\end{itemize}
	
	\paragraph{obl~\ra~advmod}
	
	\begin{itemize}
		\item \textit{Then all the girls became jealous and stopped \sout{to laugh} laughing.}
		\item \textit{I never had a chance \sout{for staying} to stay in a tent.}
	\end{itemize}
	
	\paragraph{det~\ra~nmod:poss}
	
	\begin{itemize}
		\item \textit{It is simple to find the way from \sout{this} your hotel to the conference.}
		\item \textit{It allowed me to worry about other things in \sout{the} my life.}
	\end{itemize}
	
	\paragraph{advmod~\ra~obl}
	
	\begin{itemize}
		\item \textit{And it is even more difficult to predict \sout{how} what the clothes of the future will look like.}
		\item \textit{My work was for the singers, so I could meet singers and talk \sout{together} with them.}
	\end{itemize}

	\FloatBarrier
	\section{Confusion Matrices}
	
	\subsection{POS and edge-labels changes in TLE}\label{sec:matrices_TLE}
	\FloatBarrier
	Distribution of edit types in terms of POS tags and edge labels are given in Tables \ref{table:pos_confusion_matrix} and \ref{table:edge_type_confusion_matrix} respectively. Automatic ones are given in \ref{table:auto_pos_type_confusion_matrix} and \ref{table:auto_edge_type_confusion_matrix}.

	\begin{table*}[ht]
		\centering
		\resizebox{\textwidth}{!}{%
\begin{tabular}{@{}lllllllllllllll@{}}
				\toprule
				& ADJ & ADP & ADV & AUX & CCONJ & DET & NOUN & NUM & PART & PRON & PROPN & PUNCT & SCONJ & VERB \\ \midrule
				ADJ & 444 & 3 & 27 & 1 & 0 & 16 & 49 & 2 & 0 & 2 & 0 & 0 & 0 & 21 \\
				ADP & 4 & 727 & 11 & 0 & 2 & 10 & 5 & 0 & 1 & 0 & 0 & 0 & 6 & 23 \\
				ADV & 23 & 6 & 270 & 14 & 4 & 2 & 22 & 0 & 1 & 23 & 1 & 8 & 11 & 11 \\
				AUX & 4 & 1 & 7 & 372 & 0 & 0 & 8 & 0 & 8 & 15 & 0 & 0 & 0 & 21 \\
				CCONJ & 1 & 4 & 3 & 0 & 40 & 1 & 0 & 0 & 0 & 0 & 0 & 1 & 5 & 0 \\
				DET & 18 & 11 & 7 & 1 & 0 & 315 & 3 & 1 & 6 & 34 & 0 & 2 & 5 & 5 \\
				NOUN & 50 & 5 & 22 & 5 & 0 & 3 & 1621 & 3 & 0 & 14 & 11 & 0 & 3 & 51 \\
				NUM & 1 & 0 & 0 & 0 & 0 & 6 & 3 & 21 & 0 & 0 & 0 & 0 & 0 & 0 \\
				PART & 0 & 0 & 4 & 4 & 0 & 6 & 0 & 0 & 2 & 0 & 0 & 0 & 2 & 1 \\
				PRON & 1 & 2 & 6 & 10 & 0 & 32 & 15 & 5 & 0 & 240 & 1 & 6 & 7 & 5 \\
				PROPN & 3 & 0 & 0 & 0 & 0 & 0 & 10 & 0 & 0 & 1 & 48 & 0 & 0 & 0 \\
				PUNCT & 0 & 1 & 1 & 0 & 3 & 0 & 0 & 0 & 0 & 0 & 0 & 256 & 1 & 1 \\
				SCONJ & 1 & 5 & 12 & 2 & 6 & 1 & 0 & 0 & 5 & 3 & 0 & 1 & 64 & 2 \\
				VERB & 37 & 7 & 4 & 24 & 0 & 0 & 46 & 0 & 0 & 2 & 2 & 0 & 0 & 1818 \\ \bottomrule
			\end{tabular}
		}
		\caption{\centering POS confusion matrix for replacement edits in absolute counts for TLE dataset with manual UD.}
		\label{table:pos_confusion_matrix}
	\end{table*}
	
	\begin{table*}[ht]
		\centering
		\resizebox{\textwidth}{!}{%
			\begin{tabular}{@{}lllllllllllllllllllllllllllllll@{}}
				\toprule
				& acl & acl:relcl & advcl & advmod & amod & appos & aux & aux:pass & case & cc & ccomp & compound & conj & cop & csubj & det & expl & mark & nmod & nmod:poss & nsubj & nsubj:pass & nummod & obj & obl & obl:tmod & parataxis & punct & root & xcomp \\ \midrule
				acl & 58 & 6 & 4 & 0 & 1 & 0 & 0 & 0 & 4 & 0 & 0 & 0 & 0 & 0 & 0 & 0 & 0 & 0 & 3 & 0 & 0 & 0 & 0 & 0 & 0 & 0 & 0 & 0 & 0 & 4 \\
				acl:relcl & 1 & 140 & 3 & 1 & 0 & 0 & 1 & 0 & 0 & 0 & 0 & 0 & 0 & 2 & 0 & 0 & 0 & 0 & 0 & 0 & 1 & 0 & 0 & 0 & 0 & 0 & 0 & 0 & 1 & 0 \\
				advcl & 7 & 3 & 321 & 3 & 0 & 0 & 4 & 0 & 3 & 0 & 3 & 0 & 2 & 2 & 1 & 0 & 0 & 0 & 0 & 0 & 1 & 0 & 0 & 1 & 11 & 0 & 4 & 0 & 4 & 17 \\
				advmod & 2 & 0 & 3 & 270 & 6 & 0 & 13 & 1 & 4 & 5 & 3 & 0 & 1 & 0 & 1 & 8 & 1 & 8 & 1 & 0 & 10 & 1 & 0 & 7 & 24 & 0 & 0 & 6 & 3 & 7 \\
				amod & 0 & 0 & 0 & 5 & 214 & 0 & 0 & 0 & 2 & 0 & 1 & 5 & 4 & 0 & 0 & 15 & 0 & 0 & 1 & 1 & 2 & 0 & 2 & 7 & 3 & 0 & 0 & 0 & 0 & 1 \\
				appos & 0 & 0 & 0 & 0 & 0 & 14 & 0 & 0 & 0 & 0 & 0 & 1 & 0 & 0 & 0 & 0 & 0 & 0 & 0 & 0 & 0 & 0 & 0 & 0 & 0 & 0 & 1 & 0 & 0 & 0 \\
				aux & 0 & 1 & 1 & 2 & 1 & 0 & 223 & 3 & 0 & 0 & 1 & 0 & 3 & 23 & 0 & 0 & 1 & 3 & 0 & 0 & 14 & 1 & 0 & 0 & 0 & 0 & 0 & 0 & 4 & 0 \\
\bottomrule
			\end{tabular}
		}
		\caption{\centering Edge-type confusion matrix for replacement edits in absolute counts for TLE dataset with manual UD.}
		\label{table:edge_type_confusion_matrix}
	\end{table*}

	\begin{table*}[htb] \resizebox{\textwidth}{!}{%
			\begin{tabular}{lllllllllllllllllll}
				& ADJ & ADP & ADV & AUX & CCONJ & DET & INTJ & NOUN & NUM & PART & PRON & PROPN & PUNCT & SCONJ & SYM & VERB & X & Other \\
				ADJ & 3907 & 24 & 61 & 2 & 2 & 192 & 1 & 78 & 4 & 0 & 15 & 4 & 11 & 4 & 0 & 48 & 0 & 0 \\
				ADP & 24 & 5998 & 69 & 14 & 11 & 24 & 2 & 63 & 5 & 26 & 39 & 2 & 31 & 82 & 0 & 232 & 0 & 0 \\
				ADV & 69 & 67 & 3952 & 41 & 10 & 48 & 4 & 44 & 0 & 16 & 58 & 17 & 37 & 29 & 0 & 52 & 0 & 0 \\
				AUX & 3 & 10 & 27 & 5046 & 4 & 7 & 0 & 16 & 0 & 4 & 72 & 4 & 12 & 10 & 0 & 143 & 0 & 0 \\
				CCONJ & 2 & 8 & 8 & 0 & 2428 & 3 & 0 & 12 & 0 & 0 & 7 & 2 & 14 & 6 & 0 & 4 & 0 & 0 \\
				DET & 33 & 172 & 39 & 26 & 18 & 5451 & 0 & 33 & 6 & 10 & 78 & 0 & 19 & 30 & 0 & 76 & 0 & 0 \\
				INTJ & 2 & 2 & 2 & 0 & 0 & 1 & 50 & 3 & 0 & 1 & 1 & 0 & 2 & 0 & 0 & 0 & 0 & 0 \\
				NOUN & 117 & 20 & 33 & 9 & 5 & 73 & 0 & 11288 & 8 & 6 & 19 & 37 & 17 & 8 & 0 & 146 & 0 & 0 \\
				NUM & 17 & 7 & 2 & 2 & 0 & 11 & 0 & 7 & 573 & 0 & 2 & 1 & 1 & 0 & 0 & 3 & 0 & 0 \\
				PART & 0 & 24 & 9 & 10 & 1 & 8 & 0 & 4 & 0 & 1971 & 14 & 0 & 1 & 22 & 0 & 23 & 0 & 0 \\
				PRON & 7 & 46 & 28 & 23 & 10 & 54 & 0 & 23 & 6 & 0 & 9647 & 2 & 132 & 51 & 0 & 45 & 0 & 0 \\
				PROPN & 5 & 3 & 2 & 0 & 0 & 36 & 2 & 44 & 0 & 0 & 1 & 1284 & 4 & 0 & 0 & 5 & 0 & 0 \\
				PUNCT & 9 & 8 & 10 & 8 & 11 & 6 & 0 & 57 & 0 & 1 & 18 & 3 & 1731 & 7 & 1 & 22 & 1 & 0 \\
				SCONJ & 7 & 75 & 28 & 3 & 11 & 9 & 0 & 15 & 0 & 13 & 51 & 1 & 24 & 1846 & 0 & 28 & 1 & 0 \\
				SYM & 0 & 0 & 0 & 0 & 0 & 0 & 0 & 0 & 0 & 0 & 0 & 0 & 0 & 0 & 24 & 0 & 0 & 0 \\
				VERB & 80 & 74 & 47 & 153 & 2 & 22 & 2 & 127 & 2 & 21 & 19 & 6 & 18 & 4 & 0 & 9004 & 2 & 0 \\
				X & 1 & 1 & 0 & 0 & 0 & 3 & 0 & 3 & 0 & 0 & 0 & 1 & 0 & 0 & 0 & 1 & 15 & 0
			\end{tabular}
		}
		\caption{\centering POS-type confusion matrix for replacement edits in absolute counts for TLE dataset with automatic UD.}
		\label{table:auto_pos_type_confusion_matrix}
	\end{table*}
	
	\begin{table*}[htb] \resizebox{\textwidth}{!}{%
			\begin{tabular}{lllllllllllllllllllllllllllllllllllllllllllll}
				& acl & acl:relcl & advcl & advmod & amod & appos & aux & aux:pass & case & cc & cc:preconj & ccomp & compound & compound:prt & conj & cop & csubj & det & det:predet & discourse & expl & fixed & flat & goeswith & iobj & list & mark & nmod & nmod:npmod & nmod:poss & nmod:tmod & nsubj & nsubj:pass & nummod & obj & obl & obl:npmod & obl:tmod & parataxis & punct & reparandum & vocative & xcomp & Other \\
				acl & 292 & 4 & 5 & 0 & 2 & 0 & 1 & 0 & 0 & 1 & 0 & 4 & 0 & 0 & 1 & 0 & 3 & 0 & 0 & 0 & 0 & 0 & 0 & 0 & 0 & 0 & 0 & 6 & 0 & 0 & 0 & 16 & 1 & 0 & 2 & 0 & 0 & 0 & 2 & 2 & 0 & 0 & 0 & 82 \\
				acl:relcl & 10 & 755 & 8 & 2 & 0 & 0 & 0 & 0 & 0 & 0 & 0 & 2 & 1 & 0 & 3 & 2 & 0 & 0 & 0 & 1 & 0 & 0 & 0 & 0 & 0 & 0 & 5 & 3 & 0 & 0 & 0 & 8 & 2 & 0 & 12 & 18 & 1 & 10 & 7 & 0 & 0 & 0 & 3 & 153 \\
				advcl & 6 & 6 & 1354 & 8 & 2 & 0 & 3 & 0 & 0 & 0 & 0 & 27 & 0 & 0 & 16 & 2 & 11 & 3 & 0 & 0 & 0 & 0 & 0 & 0 & 0 & 0 & 2 & 1 & 0 & 0 & 0 & 3 & 1 & 0 & 4 & 15 & 0 & 1 & 11 & 5 & 0 & 1 & 31 & 391 \\
				advmod & 1 & 5 & 12 & 3481 & 21 & 2 & 21 & 4 & 29 & 7 & 0 & 8 & 4 & 4 & 8 & 20 & 1 & 17 & 1 & 3 & 1 & 3 & 0 & 1 & 0 & 0 & 24 & 5 & 0 & 2 & 0 & 30 & 4 & 0 & 35 & 41 & 0 & 2 & 3 & 7 & 0 & 0 & 13 & 475 \\
				amod & 2 & 1 & 3 & 22 & 2345 & 0 & 0 & 0 & 23 & 1 & 1 & 0 & 22 & 0 & 3 & 1 & 0 & 174 & 3 & 0 & 0 & 0 & 0 & 0 & 0 & 0 & 1 & 2 & 0 & 4 & 0 & 10 & 0 & 4 & 21 & 9 & 2 & 1 & 1 & 4 & 0 & 0 & 1 & 121 \\
				appos & 0 & 0 & 0 & 1 & 1 & 30 & 0 & 0 & 0 & 0 & 0 & 0 & 2 & 0 & 7 & 0 & 0 & 1 & 0 & 0 & 0 & 0 & 0 & 0 & 0 & 0 & 0 & 1 & 1 & 0 & 0 & 2 & 0 & 0 & 3 & 0 & 0 & 0 & 2 & 0 & 0 & 0 & 0 & 23 \\
				aux & 2 & 3 & 4 & 16 & 1 & 0 & 2604 & 17 & 1 & 3 & 0 & 1 & 2 & 0 & 1 & 19 & 0 & 0 & 0 & 0 & 1 & 0 & 0 & 0 & 0 & 0 & 4 & 0 & 0 & 0 & 0 & 39 & 3 & 0 & 12 & 2 & 1 & 0 & 0 & 7 & 0 & 0 & 2 & 110 \\
\end{tabular}
		}
		\caption{\centering Edge-type confusion matrix for replacement edits in absolute counts for TLE dataset with automatic UD.}
		\label{table:auto_edge_type_confusion_matrix}
	\end{table*}

	\FloatBarrier
	\subsection{Output Matrices} 
	\label{sec:out_matrices}
	\FloatBarrier
	
	We supply the reader with matrices of the systems of the BEA2019 shared task that shared their outputs on dev set (avoiding over analysing the test set).
	
	\begin{table*}[htb]
		\resizebox{\textwidth}{!}{%
			\begin{tabular}{llllll|ll}
				& UEDIN-MS & KAKAO\&BRAIN & SHUYAO & CAMB\_CUED & TOHOKU & GOLD & Average \\
				INTJ & 0.86 & 0.86 & 0.71 & 1.00 & 0.43 & 472 & 0.77 \\
				PRON & 0.63 & 0.74 & 0.70 & 0.75 & 0.44 & 233 & 0.65 \\
				PROPN & 0.54 & 0.44 & 0.96 & 0.60 & 0.64 & 405 & 0.64 \\
				ADJ & 0.61 & 0.50 & 0.63 & 0.66 & 0.45 & 7 & 0.57 \\
				DET & 0.61 & 0.40 & 0.56 & 0.68 & 0.52 & 166 & 0.55 \\
				PART & 0.58 & 0.45 & 0.60 & 0.64 & 0.43 & 158 & 0.54 \\
				ADV & 0.58 & 0.51 & 0.57 & 0.66 & 0.33 & 114 & 0.53 \\
				AUX & 0.49 & 0.37 & 0.53 & 0.60 & 0.46 & 462 & 0.49 \\
				NOUN & 0.56 & 0.35 & 0.53 & 0.53 & 0.44 & 310 & 0.48 \\
				CCONJ & 0.45 & 0.61 & 0.42 & 0.59 & 0.24 & 568 & 0.46 \\
				SCONJ & 0.49 & 0.48 & 0.49 & 0.54 & 0.29 & 48 & 0.46 \\
				VERB & 0.50 & 0.36 & 0.50 & 0.53 & 0.37 & 191 & 0.45 \\
				ADP & 0.50 & 0.36 & 0.46 & 0.52 & 0.39 & 584 & 0.45 \\
				NUM & 0.46 & 0.48 & 0.42 & 0.50 & 0.31 & 618 & 0.43 \\
				SYM & 0.67 & 0.00 & 0.42 & 0.42 & 0.67 & 12 & 0.43 \\
				PUNCT & 0.61 & 0.21 & 0.40 & 0.40 & 0.38 & 461 & 0.40
			\end{tabular}%
		}
		\caption{Upper bound on recall per system. The gold amount which is considered the 100\% per row and the average per line are reported on the two rightmost columns. The order of the table is descending on the average.\label{tab:average_out_pos}}
	\end{table*}
	
	\begin{table*}[htb]
		\resizebox{\textwidth}{!}{%
			\begin{tabular}{lllllll}
				& UEDIN-MS & KAKAO\&BRAIN & SHUYAO & CAMB-CUED & TOHOKU & GOLD \\
				ADJ   & 283      & 231          & 292    & 305       & 210    & 462  \\
				ADP   & 232      & 166          & 214    & 239       & 181    & 461  \\
				ADV   & 273      & 242          & 268    & 313       & 155    & 472  \\
				AUX   & 153      & 115          & 165    & 187       & 142    & 310  \\
				CCONJ & 71       & 96           & 67     & 94        & 38     & 158  \\
				DET   & 348      & 229          & 316    & 387       & 294    & 568  \\
				INTJ  & 6        & 6            & 5      & 7         & 3      & 7    \\
				NOUN  & 346      & 219          & 328    & 330       & 273    & 618  \\
				NUM   & 22       & 23           & 20     & 24        & 15     & 48   \\
				PART  & 97       & 74           & 99     & 107       & 71     & 166  \\
				PRON  & 367      & 433          & 409    & 437       & 258    & 584  \\
				PROPN & 62       & 50           & 109    & 68        & 73     & 114  \\
				PUNCT & 116      & 40           & 76     & 77        & 73     & 191  \\
				SCONJ & 114      & 111          & 115    & 126       & 67     & 233  \\
				SYM   & 8        & 0            & 5      & 5         & 8      & 12   \\
				VERB  & 202      & 145          & 201    & 216       & 149    & 405  \\
				X     & 13       & 9            & 13     & 15        & 16     & 18  
			\end{tabular}%
		}
		\caption{ Amount of SEs per source POS for each of the outputs and the gold standard.
			\label{tab:pos-outputs}}
	\end{table*}
	
	\begin{table*}[htb] \resizebox{\textwidth}{!}{%
			\begin{tabular}{lllllllllllllllllll}
				& ADJ & ADP & ADV & AUX & CCONJ & DET & INTJ & NOUN & NUM & PART & PRON & PROPN & PUNCT & SCONJ & SYM & VERB & X & Other \\
				ADJ & 6281 & 20 & 21 & 4 & 3 & 109 & 0 & 46 & 0 & 0 & 9 & 8 & 28 & 2 & 0 & 33 & 0 & 0 \\
				ADP & 17 & 7649 & 10 & 10 & 0 & 3 & 0 & 25 & 1 & 22 & 16 & 1 & 52 & 35 & 0 & 40 & 0 & 0 \\
				ADV & 28 & 18 & 4650 & 23 & 5 & 17 & 8 & 17 & 0 & 3 & 25 & 2 & 95 & 14 & 0 & 18 & 0 & 0 \\
				AUX & 2 & 1 & 8 & 5680 & 4 & 2 & 0 & 19 & 0 & 10 & 51 & 2 & 11 & 2 & 0 & 41 & 0 & 0 \\
				CCONJ & 0 & 2 & 2 & 1 & 2946 & 2 & 0 & 9 & 0 & 0 & 2 & 1 & 51 & 0 & 0 & 1 & 0 & 0 \\
				DET & 15 & 112 & 13 & 13 & 8 & 7175 & 1 & 11 & 1 & 8 & 31 & 2 & 61 & 13 & 0 & 58 & 1 & 0 \\
				INTJ & 0 & 1 & 0 & 0 & 0 & 0 & 92 & 1 & 0 & 0 & 0 & 1 & 2 & 0 & 0 & 1 & 0 & 0 \\
				NOUN & 67 & 7 & 19 & 10 & 5 & 35 & 0 & 16050 & 2 & 3 & 18 & 33 & 55 & 5 & 0 & 87 & 0 & 0 \\
				NUM & 3 & 1 & 2 & 1 & 0 & 3 & 0 & 6 & 628 & 0 & 0 & 0 & 4 & 1 & 0 & 1 & 0 & 0 \\
				PART & 2 & 10 & 6 & 9 & 1 & 2 & 0 & 9 & 0 & 2523 & 13 & 1 & 6 & 20 & 0 & 18 & 0 & 0 \\
				PRON & 1 & 13 & 16 & 17 & 6 & 12 & 0 & 25 & 0 & 2 & 8016 & 1 & 235 & 14 & 0 & 25 & 0 & 0 \\
				PROPN & 14 & 1 & 0 & 0 & 1 & 16 & 0 & 19 & 0 & 1 & 1 & 1926 & 6 & 0 & 0 & 3 & 0 & 0 \\
				PUNCT & 7 & 8 & 10 & 8 & 8 & 5 & 1 & 29 & 1 & 6 & 5 & 6 & 8679 & 11 & 0 & 11 & 0 & 0 \\
				SCONJ & 0 & 25 & 7 & 0 & 1 & 2 & 0 & 6 & 0 & 17 & 12 & 1 & 31 & 1882 & 0 & 12 & 0 & 0 \\
				SYM & 1 & 0 & 0 & 0 & 0 & 0 & 0 & 1 & 0 & 1 & 1 & 0 & 3 & 0 & 45 & 1 & 0 & 0 \\
				VERB & 21 & 9 & 9 & 44 & 3 & 7 & 1 & 67 & 0 & 5 & 7 & 5 & 20 & 3 & 0 & 9973 & 1 & 0 \\
				X & 0 & 0 & 2 & 0 & 0 & 2 & 0 & 2 & 0 & 0 & 2 & 0 & 2 & 0 & 0 & 3 & 65 & 0
			\end{tabular}
		}
		\caption{UEDIN-MS POS confusion matrix on W\&I development.}
	\end{table*}
	
	\begin{table*}[htb] \resizebox{\textwidth}{!}{%
			\begin{tabular}{lllllllllllllllllll}
				& ADJ & ADP & ADV & AUX & CCONJ & DET & INTJ & NOUN & NUM & PART & PRON & PROPN & PUNCT & SCONJ & SYM & VERB & X & Other \\
				ADJ & 6354 & 16 & 18 & 3 & 1 & 92 & 0 & 37 & 0 & 1 & 2 & 6 & 13 & 0 & 0 & 21 & 0 & 0 \\
				ADP & 19 & 7700 & 14 & 6 & 0 & 4 & 0 & 30 & 1 & 12 & 8 & 4 & 11 & 25 & 0 & 47 & 0 & 0 \\
				ADV & 15 & 12 & 4768 & 19 & 4 & 13 & 4 & 15 & 0 & 2 & 12 & 3 & 25 & 11 & 0 & 20 & 0 & 0 \\
				AUX & 1 & 2 & 10 & 5691 & 3 & 1 & 0 & 22 & 0 & 12 & 41 & 0 & 6 & 1 & 0 & 43 & 0 & 0 \\
				CCONJ & 2 & 0 & 1 & 1 & 2979 & 2 & 0 & 6 & 0 & 1 & 2 & 1 & 16 & 0 & 1 & 5 & 0 & 0 \\
				DET & 13 & 93 & 21 & 10 & 9 & 7229 & 0 & 12 & 0 & 5 & 17 & 1 & 45 & 10 & 0 & 57 & 1 & 0 \\
				INTJ & 0 & 0 & 0 & 0 & 0 & 0 & 95 & 0 & 0 & 0 & 2 & 0 & 1 & 0 & 0 & 0 & 0 & 0 \\
				NOUN & 46 & 6 & 20 & 2 & 5 & 33 & 0 & 16123 & 3 & 3 & 22 & 21 & 31 & 5 & 0 & 76 & 0 & 0 \\
				NUM & 1 & 0 & 0 & 0 & 1 & 3 & 0 & 5 & 635 & 0 & 1 & 0 & 1 & 1 & 0 & 2 & 0 & 0 \\
				PART & 2 & 7 & 4 & 9 & 0 & 0 & 0 & 11 & 0 & 2549 & 11 & 1 & 2 & 13 & 0 & 11 & 0 & 0 \\
				PRON & 2 & 11 & 13 & 14 & 7 & 10 & 0 & 28 & 0 & 2 & 8125 & 2 & 128 & 13 & 0 & 28 & 0 & 0 \\
				PROPN & 19 & 5 & 0 & 1 & 0 & 13 & 0 & 24 & 0 & 0 & 3 & 1915 & 4 & 0 & 0 & 3 & 1 & 0 \\
				PUNCT & 11 & 2 & 8 & 1 & 5 & 2 & 1 & 22 & 0 & 0 & 4 & 2 & 8722 & 4 & 1 & 10 & 0 & 0 \\
				SCONJ & 0 & 12 & 9 & 4 & 1 & 3 & 0 & 5 & 0 & 9 & 7 & 0 & 8 & 1929 & 0 & 9 & 0 & 0 \\
				SYM & 2 & 0 & 0 & 0 & 0 & 0 & 0 & 2 & 0 & 0 & 0 & 0 & 3 & 0 & 45 & 1 & 0 & 0 \\
				VERB & 17 & 7 & 7 & 30 & 3 & 6 & 3 & 44 & 0 & 7 & 9 & 2 & 11 & 2 & 0 & 10026 & 1 & 0 \\
				X & 1 & 0 & 2 & 1 & 0 & 1 & 0 & 6 & 0 & 0 & 1 & 2 & 1 & 0 & 0 & 1 & 62 & 0
			\end{tabular}
		}
		\caption{AIP-TOHOKU POS confusion matrix on W\&I development.}
	\end{table*}
	
	\begin{table*}[htb] \resizebox{\textwidth}{!}{%
			\begin{tabular}{lllllllllllllllllll}
				& ADJ & ADP & ADV & AUX & CCONJ & DET & INTJ & NOUN & NUM & PART & PRON & PROPN & PUNCT & SCONJ & SYM & VERB & X & Other \\
				ADJ & 6333 & 9 & 16 & 2 & 0 & 98 & 0 & 41 & 0 & 0 & 5 & 7 & 26 & 0 & 0 & 27 & 0 & 0 \\
				ADP & 10 & 7715 & 7 & 6 & 1 & 4 & 0 & 14 & 0 & 10 & 8 & 1 & 39 & 33 & 0 & 33 & 0 & 0 \\
				ADV & 20 & 16 & 4681 & 13 & 1 & 12 & 10 & 6 & 0 & 2 & 12 & 4 & 124 & 9 & 0 & 13 & 0 & 0 \\
				AUX & 0 & 0 & 2 & 5718 & 1 & 1 & 0 & 11 & 0 & 9 & 32 & 0 & 11 & 4 & 0 & 44 & 0 & 0 \\
				CCONJ & 0 & 0 & 2 & 0 & 2921 & 2 & 0 & 3 & 0 & 0 & 0 & 2 & 86 & 0 & 0 & 1 & 0 & 0 \\
				DET & 12 & 59 & 4 & 2 & 3 & 7294 & 1 & 6 & 0 & 5 & 21 & 0 & 81 & 10 & 0 & 25 & 0 & 0 \\
				INTJ & 0 & 0 & 0 & 0 & 0 & 0 & 92 & 0 & 0 & 0 & 2 & 1 & 2 & 0 & 0 & 1 & 0 & 0 \\
				NOUN & 31 & 3 & 11 & 2 & 3 & 23 & 0 & 16177 & 3 & 0 & 11 & 17 & 48 & 4 & 0 & 63 & 0 & 0 \\
				NUM & 3 & 1 & 0 & 1 & 0 & 4 & 0 & 5 & 627 & 0 & 0 & 0 & 9 & 0 & 0 & 0 & 0 & 0 \\
				PART & 1 & 9 & 3 & 13 & 0 & 0 & 0 & 8 & 0 & 2546 & 8 & 0 & 4 & 18 & 0 & 10 & 0 & 0 \\
				PRON & 2 & 12 & 6 & 12 & 0 & 8 & 0 & 17 & 0 & 1 & 7950 & 1 & 346 & 11 & 0 & 17 & 0 & 0 \\
				PROPN & 2 & 1 & 0 & 0 & 0 & 12 & 0 & 20 & 0 & 0 & 1 & 1938 & 13 & 0 & 0 & 1 & 0 & 0 \\
				PUNCT & 1 & 6 & 3 & 2 & 4 & 0 & 0 & 10 & 0 & 1 & 3 & 0 & 8755 & 5 & 0 & 5 & 0 & 0 \\
				SCONJ & 0 & 24 & 7 & 0 & 0 & 1 & 0 & 4 & 0 & 18 & 10 & 0 & 37 & 1885 & 0 & 10 & 0 & 0 \\
				SYM & 0 & 0 & 0 & 0 & 0 & 0 & 0 & 0 & 0 & 0 & 0 & 0 & 0 & 0 & 53 & 0 & 0 & 0 \\
				VERB & 16 & 4 & 5 & 27 & 1 & 10 & 1 & 47 & 0 & 1 & 5 & 3 & 21 & 4 & 0 & 10030 & 0 & 0 \\
				X & 0 & 0 & 1 & 1 & 0 & 1 & 0 & 4 & 0 & 0 & 1 & 0 & 0 & 0 & 0 & 1 & 69 & 0
			\end{tabular}
		}
		\caption{Kakao\&Brain POS confusion matrix on W\&I development.}
	\end{table*}
	
	\begin{table*}[htb] \resizebox{\textwidth}{!}{%
			\begin{tabular}{lllllllllllllllllll}
				& ADJ & ADP & ADV & AUX & CCONJ & DET & INTJ & NOUN & NUM & PART & PRON & PROPN & PUNCT & SCONJ & SYM & VERB & X & Other \\
				ADJ & 6272 & 20 & 30 & 1 & 0 & 100 & 0 & 57 & 3 & 0 & 5 & 9 & 33 & 0 & 0 & 34 & 0 & 0 \\
				ADP & 13 & 7667 & 11 & 6 & 1 & 3 & 0 & 20 & 0 & 17 & 14 & 2 & 49 & 37 & 0 & 41 & 0 & 0 \\
				ADV & 28 & 18 & 4655 & 15 & 6 & 15 & 6 & 16 & 0 & 1 & 16 & 2 & 115 & 12 & 0 & 18 & 0 & 0 \\
				AUX & 2 & 2 & 10 & 5668 & 1 & 2 & 0 & 23 & 0 & 9 & 48 & 3 & 16 & 1 & 0 & 48 & 0 & 0 \\
				CCONJ & 0 & 1 & 1 & 1 & 2950 & 3 & 0 & 7 & 0 & 0 & 2 & 0 & 52 & 0 & 0 & 0 & 0 & 0 \\
				DET & 17 & 98 & 9 & 11 & 6 & 7207 & 2 & 7 & 0 & 6 & 25 & 1 & 81 & 9 & 0 & 44 & 0 & 0 \\
				INTJ & 0 & 0 & 0 & 0 & 0 & 0 & 93 & 0 & 0 & 0 & 0 & 3 & 2 & 0 & 0 & 0 & 0 & 0 \\
				NOUN & 64 & 9 & 20 & 4 & 6 & 26 & 0 & 16068 & 11 & 1 & 15 & 26 & 59 & 5 & 0 & 82 & 0 & 0 \\
				NUM & 1 & 1 & 2 & 0 & 0 & 2 & 0 & 6 & 630 & 0 & 0 & 0 & 7 & 0 & 0 & 1 & 0 & 0 \\
				PART & 2 & 11 & 6 & 10 & 1 & 2 & 0 & 12 & 0 & 2521 & 11 & 0 & 4 & 22 & 0 & 18 & 0 & 0 \\
				PRON & 3 & 14 & 11 & 13 & 3 & 11 & 0 & 24 & 2 & 2 & 7974 & 1 & 288 & 14 & 0 & 23 & 0 & 0 \\
				PROPN & 12 & 3 & 3 & 0 & 0 & 16 & 1 & 50 & 0 & 1 & 6 & 1879 & 10 & 0 & 0 & 6 & 1 & 0 \\
				PUNCT & 6 & 6 & 6 & 4 & 1 & 0 & 0 & 35 & 1 & 0 & 1 & 0 & 8719 & 10 & 0 & 6 & 0 & 0 \\
				SCONJ & 1 & 21 & 7 & 0 & 1 & 2 & 0 & 7 & 0 & 18 & 8 & 0 & 38 & 1881 & 0 & 12 & 0 & 0 \\
				SYM & 1 & 0 & 0 & 0 & 0 & 0 & 0 & 2 & 0 & 0 & 0 & 0 & 1 & 0 & 48 & 1 & 0 & 0 \\
				VERB & 17 & 11 & 10 & 34 & 1 & 7 & 1 & 77 & 1 & 2 & 7 & 4 & 26 & 3 & 0 & 9974 & 0 & 0 \\
				X & 1 & 0 & 2 & 1 & 0 & 1 & 0 & 3 & 0 & 1 & 1 & 0 & 2 & 0 & 0 & 1 & 65 & 0
			\end{tabular}
		}
		\caption{SHUYAO POS confusion matrix on W\&I development.}
	\end{table*}
	
	\begin{table*}[htb] \resizebox{\textwidth}{!}{%
			\begin{tabular}{lllllllllllllllllll}
				& ADJ & ADP & ADV & AUX & CCONJ & DET & INTJ & NOUN & NUM & PART & PRON & PROPN & PUNCT & SCONJ & SYM & VERB & X & Other \\
				ADJ & 6259 & 23 & 24 & 3 & 4 & 118 & 0 & 55 & 0 & 0 & 8 & 9 & 30 & 0 & 0 & 31 & 0 & 0 \\
				ADP & 15 & 7642 & 8 & 12 & 1 & 5 & 0 & 35 & 0 & 21 & 12 & 1 & 61 & 26 & 0 & 42 & 0 & 0 \\
				ADV & 23 & 19 & 4610 & 20 & 9 & 17 & 10 & 21 & 0 & 2 & 21 & 4 & 122 & 19 & 0 & 26 & 0 & 0 \\
				AUX & 3 & 1 & 14 & 5646 & 3 & 4 & 0 & 24 & 0 & 12 & 56 & 2 & 13 & 2 & 0 & 53 & 0 & 0 \\
				CCONJ & 3 & 1 & 4 & 1 & 2923 & 2 & 0 & 7 & 0 & 0 & 5 & 1 & 69 & 0 & 0 & 1 & 0 & 0 \\
				DET & 17 & 122 & 16 & 12 & 10 & 7136 & 1 & 17 & 2 & 8 & 30 & 2 & 72 & 15 & 0 & 63 & 0 & 0 \\
				INTJ & 0 & 0 & 0 & 0 & 0 & 0 & 91 & 0 & 0 & 0 & 2 & 1 & 3 & 0 & 0 & 1 & 0 & 0 \\
				NOUN & 50 & 9 & 15 & 7 & 8 & 33 & 0 & 16066 & 1 & 5 & 21 & 32 & 60 & 3 & 0 & 86 & 0 & 0 \\
				NUM & 3 & 1 & 0 & 0 & 1 & 3 & 0 & 6 & 626 & 1 & 1 & 0 & 6 & 1 & 0 & 1 & 0 & 0 \\
				PART & 3 & 10 & 12 & 14 & 1 & 0 & 0 & 12 & 0 & 2513 & 9 & 0 & 3 & 24 & 0 & 19 & 0 & 0 \\
				PRON & 4 & 21 & 9 & 17 & 10 & 9 & 0 & 27 & 0 & 2 & 7946 & 3 & 293 & 16 & 0 & 26 & 0 & 0 \\
				PROPN & 10 & 1 & 0 & 0 & 0 & 16 & 0 & 32 & 1 & 0 & 1 & 1920 & 6 & 0 & 0 & 1 & 0 & 0 \\
				PUNCT & 1 & 4 & 7 & 5 & 4 & 2 & 2 & 28 & 0 & 2 & 6 & 3 & 8718 & 7 & 0 & 6 & 0 & 0 \\
				SCONJ & 1 & 20 & 8 & 1 & 1 & 2 & 0 & 6 & 0 & 13 & 12 & 0 & 51 & 1870 & 0 & 11 & 0 & 0 \\
				SYM & 0 & 0 & 0 & 0 & 0 & 0 & 0 & 1 & 0 & 0 & 0 & 0 & 3 & 0 & 48 & 1 & 0 & 0 \\
				VERB & 25 & 9 & 11 & 48 & 4 & 6 & 1 & 63 & 0 & 6 & 13 & 1 & 25 & 3 & 0 & 9959 & 1 & 0 \\
				X & 0 & 0 & 2 & 1 & 1 & 2 & 0 & 6 & 0 & 1 & 1 & 0 & 1 & 0 & 0 & 0 & 63 & 0
			\end{tabular}
		}
		\caption{CAMB-CUED POS confusion matrix on W\&I development.}
	\end{table*}
	\FloatBarrier
	\subsection{W\&I by levels}\label{sec:wi_matrices}
	\FloatBarrier
	
	We provide a separate matrix for each level of learner proficiency in the W\&I corpus and the native proficiency of LOCNESS. 
	\begin{table*}[htb]
		\resizebox{\textwidth}{!}{%
			\begin{tabular}{llllllllllllllllll}
				& ADJ & ADP & ADV & AUX & CCONJ & DET & INTJ & NOUN & NUM & PART & PRON & PROPN & PUNCT & SCONJ & SYM & VERB & X \\
				ADJ & 11594 & 70 & 190 & 11 & 11 & 442 & 2 & 456 & 7 & 5 & 36 & 65 & 123 & 12 & 0 & 225 & 0 \\
				ADP & 101 & 13880 & 167 & 35 & 33 & 70 & 7 & 233 & 12 & 103 & 82 & 14 & 266 & 193 & 1 & 388 & 0 \\
				ADV & 181 & 118 & 8493 & 119 & 78 & 107 & 28 & 173 & 5 & 23 & 134 & 18 & 361 & 80 & 1 & 147 & 0 \\
				AUX & 23 & 8 & 59 & 10907 & 9 & 16 & 1 & 111 & 1 & 39 & 324 & 14 & 75 & 8 & 1 & 513 & 0 \\
				CCONJ & 13 & 17 & 9 & 4 & 5843 & 9 & 0 & 45 & 3 & 2 & 8 & 0 & 365 & 6 & 0 & 11 & 0 \\
				DET & 82 & 522 & 105 & 115 & 66 & 12806 & 5 & 147 & 22 & 22 & 186 & 9 & 341 & 112 & 0 & 410 & 0 \\
				INTJ & 4 & 0 & 4 & 2 & 0 & 3 & 312 & 8 & 0 & 1 & 1 & 5 & 9 & 1 & 0 & 5 & 0 \\
				NOUN & 372 & 71 & 170 & 71 & 25 & 156 & 12 & 31992 & 28 & 10 & 133 & 235 & 300 & 32 & 1 & 658 & 4 \\
				NUM & 8 & 17 & 8 & 7 & 1 & 32 & 0 & 48 & 1405 & 1 & 10 & 1 & 14 & 1 & 2 & 11 & 0 \\
				PART & 14 & 79 & 51 & 81 & 11 & 11 & 1 & 63 & 1 & 4845 & 38 & 7 & 32 & 104 & 0 & 125 & 0 \\
				PRON & 24 & 94 & 68 & 81 & 48 & 78 & 1 & 165 & 6 & 23 & 19290 & 15 & 1110 & 100 & 0 & 137 & 1 \\
				PROPN & 32 & 14 & 8 & 2 & 4 & 45 & 6 & 156 & 1 & 1 & 7 & 3297 & 62 & 3 & 0 & 16 & 0 \\
				PUNCT & 21 & 22 & 26 & 9 & 24 & 8 & 5 & 140 & 3 & 11 & 44 & 11 & 17928 & 26 & 1 & 58 & 1 \\
				SCONJ & 16 & 113 & 70 & 7 & 20 & 15 & 1 & 46 & 1 & 61 & 107 & 9 & 254 & 3270 & 0 & 76 & 0 \\
				SYM & 0 & 2 & 0 & 0 & 1 & 3 & 0 & 0 & 0 & 0 & 0 & 0 & 1 & 0 & 77 & 1 & 1 \\
				VERB & 169 & 59 & 65 & 251 & 9 & 40 & 4 & 393 & 1 & 21 & 43 & 28 & 103 & 21 & 3 & 21005 & 1 \\
				X & 1 & 0 & 3 & 1 & 1 & 2 & 0 & 5 & 0 & 1 & 0 & 4 & 2 & 1 & 2 & 1 & 111
			\end{tabular}%
		}
		\caption{Matrix for POS on W\&I training set of A (lowest) level learners.}
	\end{table*}
	\begin{table*}[htb]
		\resizebox{\textwidth}{!}{%
			\begin{tabular}{llllllllllllllllll}
				& ADJ & ADP & ADV & AUX & CCONJ & DET & INTJ & NOUN & NUM & PART & PRON & PROPN & PUNCT & SCONJ & SYM & VERB & X \\
				ADJ & 16232 & 66 & 155 & 14 & 9 & 385 & 2 & 319 & 10 & 2 & 37 & 64 & 121 & 17 & 0 & 158 & 1 \\
				ADP & 80 & 19910 & 132 & 37 & 31 & 57 & 4 & 166 & 15 & 61 & 58 & 11 & 263 & 165 & 1 & 307 & 1 \\
				ADV & 142 & 96 & 11753 & 126 & 40 & 67 & 27 & 158 & 1 & 31 & 119 & 8 & 365 & 89 & 0 & 134 & 1 \\
				AUX & 13 & 6 & 61 & 14903 & 4 & 15 & 1 & 92 & 2 & 33 & 213 & 8 & 92 & 14 & 0 & 340 & 0 \\
				CCONJ & 13 & 13 & 11 & 1 & 7499 & 5 & 1 & 30 & 0 & 2 & 11 & 3 & 336 & 8 & 0 & 9 & 1 \\
				DET & 70 & 536 & 99 & 88 & 63 & 17932 & 6 & 109 & 20 & 14 & 131 & 6 & 281 & 96 & 0 & 316 & 0 \\
				INTJ & 3 & 0 & 6 & 0 & 0 & 0 & 269 & 6 & 0 & 0 & 3 & 3 & 8 & 1 & 0 & 2 & 0 \\
				NOUN & 288 & 53 & 124 & 36 & 21 & 151 & 8 & 42212 & 31 & 6 & 71 & 181 & 196 & 32 & 2 & 437 & 2 \\
				NUM & 8 & 9 & 3 & 2 & 0 & 31 & 0 & 44 & 1756 & 0 & 6 & 6 & 17 & 1 & 0 & 8 & 0 \\
				PART & 16 & 60 & 49 & 43 & 4 & 12 & 0 & 55 & 0 & 6880 & 42 & 8 & 35 & 114 & 0 & 130 & 0 \\
				PRON & 14 & 65 & 44 & 58 & 45 & 74 & 1 & 139 & 13 & 15 & 23771 & 9 & 1101 & 100 & 0 & 109 & 0 \\
				PROPN & 20 & 8 & 15 & 4 & 2 & 58 & 1 & 169 & 0 & 1 & 9 & 4854 & 59 & 5 & 0 & 11 & 1 \\
				PUNCT & 48 & 20 & 30 & 18 & 35 & 21 & 2 & 172 & 0 & 8 & 34 & 14 & 23620 & 26 & 1 & 64 & 0 \\
				SCONJ & 10 & 115 & 76 & 10 & 17 & 14 & 0 & 46 & 1 & 70 & 63 & 7 & 277 & 4720 & 0 & 34 & 0 \\
				SYM & 1 & 0 & 0 & 0 & 0 & 0 & 0 & 3 & 0 & 0 & 0 & 0 & 0 & 0 & 79 & 1 & 0 \\
				VERB & 165 & 44 & 65 & 182 & 7 & 37 & 2 & 329 & 2 & 19 & 32 & 25 & 96 & 22 & 1 & 27432 & 0 \\
				X & 0 & 0 & 3 & 0 & 0 & 2 & 0 & 8 & 0 & 1 & 2 & 4 & 5 & 0 & 2 & 4 & 143
			\end{tabular}%
		}
		\caption{Matrix for POS on W\&I training set of B (mid) level learners.}
	\end{table*}
	
	\begin{table*}[htb]
		\resizebox{\textwidth}{!}{%
			\begin{tabular}{llllllllllllllllll}
				& ADJ & ADP & ADV & AUX & CCONJ & DET & INTJ & NOUN & NUM & PART & PRON & PROPN & PUNCT & SCONJ & SYM & VERB & X \\
				ADJ & 14938 & 27 & 81 & 8 & 3 & 174 & 0 & 139 & 2 & 3 & 13 & 23 & 53 & 1 & 0 & 67 & 0 \\
				ADP & 35 & 18358 & 60 & 13 & 13 & 14 & 2 & 58 & 3 & 19 & 24 & 1 & 161 & 48 & 1 & 133 & 1 \\
				ADV & 64 & 50 & 11003 & 34 & 26 & 41 & 7 & 74 & 2 & 8 & 45 & 2 & 217 & 28 & 0 & 47 & 0 \\
				AUX & 8 & 8 & 22 & 12706 & 6 & 4 & 0 & 38 & 0 & 13 & 81 & 5 & 49 & 5 & 0 & 109 & 0 \\
				CCONJ & 1 & 12 & 4 & 6 & 6767 & 9 & 0 & 15 & 0 & 2 & 6 & 0 & 167 & 0 & 0 & 3 & 0 \\
				DET & 22 & 226 & 26 & 24 & 18 & 16736 & 0 & 32 & 2 & 4 & 49 & 5 & 142 & 19 & 0 & 125 & 1 \\
				INTJ & 1 & 1 & 4 & 0 & 0 & 0 & 216 & 2 & 0 & 0 & 0 & 1 & 2 & 0 & 0 & 0 & 0 \\
				NOUN & 116 & 32 & 29 & 14 & 5 & 63 & 2 & 37581 & 8 & 2 & 38 & 52 & 91 & 4 & 1 & 172 & 2 \\
				NUM & 10 & 1 & 3 & 1 & 0 & 9 & 0 & 18 & 1501 & 0 & 7 & 0 & 6 & 0 & 1 & 5 & 1 \\
				PART & 3 & 18 & 26 & 19 & 4 & 2 & 0 & 31 & 2 & 6167 & 21 & 5 & 18 & 57 & 0 & 29 & 0 \\
				PRON & 8 & 30 & 18 & 31 & 4 & 20 & 0 & 32 & 2 & 0 & 20363 & 3 & 574 & 32 & 0 & 37 & 0 \\
				PROPN & 14 & 2 & 7 & 1 & 1 & 25 & 0 & 50 & 0 & 1 & 1 & 4511 & 52 & 0 & 0 & 5 & 0 \\
				PUNCT & 25 & 9 & 15 & 5 & 11 & 6 & 2 & 88 & 1 & 5 & 20 & 3 & 21344 & 7 & 0 & 14 & 0 \\
				SCONJ & 8 & 59 & 34 & 1 & 5 & 10 & 1 & 19 & 0 & 20 & 32 & 5 & 144 & 4239 & 0 & 16 & 0 \\
				SYM & 1 & 1 & 1 & 0 & 1 & 1 & 1 & 4 & 0 & 0 & 0 & 0 & 9 & 0 & 84 & 0 & 0 \\
				VERB & 58 & 12 & 18 & 66 & 2 & 13 & 1 & 148 & 2 & 3 & 11 & 4 & 75 & 3 & 0 & 24299 & 2 \\
				X & 0 & 0 & 1 & 0 & 0 & 2 & 0 & 1 & 0 & 0 & 0 & 2 & 1 & 0 & 0 & 0 & 120
			\end{tabular}%
		}
		\caption{Matrix for POS on W\&I training set of C (highest) level learners.}
	\end{table*}
	\begin{table*}[htb]
		\resizebox{\textwidth}{!}{%
			\begin{tabular}{llllllllllllllllll}
				& ADJ & ADP & ADV & AUX & CCONJ & DET & INTJ & NOUN & NUM & PART & PRON & PROPN & PUNCT & SCONJ & SYM & VERB & X \\
				ADJ & 1753 & 1 & 5 & 0 & 0 & 3 & 0 & 18 & 0 & 0 & 1 & 4 & 10 & 0 & 0 & 8 & 0 \\
				ADP & 4 & 2228 & 5 & 0 & 1 & 1 & 0 & 9 & 0 & 3 & 1 & 0 & 20 & 6 & 0 & 5 & 0 \\
				ADV & 6 & 6 & 1264 & 2 & 1 & 0 & 0 & 7 & 0 & 0 & 1 & 0 & 41 & 2 & 0 & 0 & 1 \\
				AUX & 0 & 0 & 0 & 1720 & 0 & 0 & 0 & 4 & 0 & 1 & 2 & 0 & 11 & 0 & 0 & 5 & 0 \\
				CCONJ & 0 & 2 & 0 & 0 & 763 & 0 & 0 & 2 & 0 & 0 & 0 & 0 & 18 & 1 & 0 & 0 & 0 \\
				DET & 3 & 17 & 2 & 4 & 3 & 2138 & 0 & 0 & 0 & 0 & 2 & 0 & 27 & 2 & 0 & 3 & 0 \\
				INTJ & 0 & 0 & 0 & 0 & 0 & 0 & 9 & 0 & 0 & 0 & 0 & 0 & 0 & 0 & 0 & 0 & 0 \\
				NOUN & 10 & 2 & 7 & 0 & 0 & 2 & 0 & 4554 & 1 & 0 & 1 & 8 & 24 & 3 & 1 & 19 & 0 \\
				NUM & 0 & 0 & 0 & 0 & 0 & 0 & 0 & 1 & 153 & 0 & 0 & 0 & 4 & 0 & 2 & 0 & 1 \\
				PART & 0 & 2 & 0 & 1 & 0 & 0 & 0 & 1 & 0 & 643 & 2 & 0 & 4 & 0 & 0 & 0 & 0 \\
				PRON & 0 & 1 & 2 & 1 & 2 & 0 & 0 & 1 & 2 & 0 & 1422 & 0 & 55 & 2 & 0 & 4 & 0 \\
				PROPN & 5 & 0 & 0 & 0 & 0 & 1 & 0 & 11 & 0 & 0 & 0 & 568 & 1 & 0 & 0 & 1 & 0 \\
				PUNCT & 4 & 3 & 2 & 1 & 0 & 1 & 0 & 22 & 1 & 1 & 1 & 1 & 2304 & 3 & 0 & 4 & 0 \\
				SCONJ & 0 & 3 & 3 & 0 & 0 & 0 & 0 & 1 & 0 & 1 & 4 & 0 & 16 & 504 & 0 & 3 & 0 \\
				SYM & 2 & 0 & 0 & 0 & 0 & 0 & 0 & 3 & 0 & 0 & 0 & 0 & 2 & 0 & 12 & 1 & 1 \\
				VERB & 2 & 1 & 2 & 4 & 0 & 0 & 0 & 23 & 0 & 0 & 1 & 4 & 12 & 0 & 0 & 2459 & 0 \\
				X & 0 & 0 & 0 & 0 & 0 & 1 & 0 & 1 & 0 & 0 & 0 & 0 & 2 & 0 & 1 & 0 & 19
			\end{tabular}%
		}
		\caption{Matrix for POS on LOCNESS training set of N (native) writers.}
	\end{table*}
	
	\FloatBarrier
	\subsection{Comparison to NUCLE}
	Confusion matrix between SeRCL and NUCLE types over the NUCLE train set is given in \ref{tab:m2_stx}. SeRCL types with less than 30 occurrences are omitted. Aggregations are find in the rightmost columns (and do not include -rloc)
	\begin{table*}[htb]
		\resizebox{\linewidth}{!}{%
			\begin{tabular}{@{}llllllllllllllllllllllllllllllllll@{}}
				{ }                          & { Wci} & { Pref} & { Prep} & { ArtOrDet} & { Vm} & { Others} & { Wtone} & { Smod} & { Mec} & { Pform} & { Trans} & { Cit} & { Wa} & { Sfrag} & { WOinc} & { Wform} & { Ssub} & { Npos} & { V0}  & { SVA} & { Srun} & { Nn}  & { Vform} & { Spar} & { Um}  & { WOadv} & { Vt}  & { max}  & { 2ns} & { 3rd} & { sum}  & { max/sum}      & { top3/sum}     \\
				{ None-\textgreater{}DET}    & { 27}  & { 6}    & { 0}    & { 3434}     & { 0}  & { 33}     & { 0}     & { 0}    & { 10}  & { 0}     & { 7}     & { 0}   & { 0}  & { 1}     & { 0}     & { 1}     & { 1}    & { 0}    & { 0}   & { 12}  & { 0}    & { 23}  & { 1}     & { 7}    & { 2}   & { 1}     & { 15}  & { 3434} & { 33}  & { 27}  & { 3581} & { 0.958950014}  & { 0.9757051103} \\
				{ CCONJ-\textgreater{}ADV}   & { 1}   & { 0}    & { 0}    & { 0}        & { 0}  & { 1}      & { 0}     & { 0}    & { 1}   & { 0}     & { 55}    & { 0}   & { 0}  & { 0}     & { 0}     & { 0}     & { 0}    & { 0}    & { 0}   & { 0}   & { 0}    & { 0}   & { 0}     & { 0}    & { 0}   & { 0}     & { 0}   & { 55}   & { 1}   & { 1}   & { 58}   & { 0.9482758621} & { 0.9827586207} \\
				{ DET-\textgreater{}None}    & { 6}   & { 4}    & { 0}    & { 1998}     & { 1}  & { 1}      & { 2}     & { 0}    & { 7}   & { 2}     & { 1}     & { 0}   & { 0}  & { 1}     & { 1}     & { 1}     & { 0}    & { 0}    & { 0}   & { 8}   & { 2}    & { 66}  & { 1}     & { 1}    & { 2}   & { 1}     & { 5}   & { 1998} & { 66}  & { 8}   & { 2111} & { 0.9464708669} & { 0.9815253434} \\
				{ ADP-\textgreater{}None}    & { 20}  & { 0}    & { 610}  & { 2}        & { 0}  & { 1}      & { 1}     & { 1}    & { 3}   & { 0}     & { 17}    & { 2}   & { 0}  & { 1}     & { 3}     & { 1}     & { 0}    & { 0}    & { 0}   & { 3}   & { 0}    & { 0}   & { 0}     & { 4}    & { 2}   & { 0}     & { 3}   & { 610}  & { 20}  & { 17}  & { 674}  & { 0.9050445104} & { 0.9599406528} \\
\end{tabular}%
		}
		\caption{NUCLE and SerCL confusion matrix \label{tab:m2_stx}}
	\end{table*}
	
	\FloatBarrier
	\subsection{Morphological feature-changes in edits}\label{sec:matrices_morph}
	\subsubsection{Agreement features on adjectives}
	\FloatBarrier
	
	We provide  the morphological aware matrices, delving into finer details than the POS tag and hence allowing for inner classifications of the cases were the POS tag was kept but its characteristics were not.
	\begin{table*}[ht]
		\begin{tabular}{lllllll}
			\hline
			& \textbf{Acc} & \textbf{Dat} & \textbf{Gen} & \textbf{Ins} & \textbf{Loc} & \textbf{Nom} \\
			\textbf{Acc} & 0 & 12 & 27 & 9 & 12 & 19 \\
			\textbf{Dat} & 2 & 0 & 6 & 2 & 1 & 2 \\
			\textbf{Gen} & 19 & 8 & 0 & 13 & 6 & 18 \\
			\textbf{Ins} & 4 & 5 & 6 & 0 & 5 & 8 \\
			\textbf{Loc} & 22 & 3 & 9 & 8 & 0 & 2 \\
			\textbf{Nom} & 32 & 5 & 30 & 48 & 3 & 0 \\ \hline
		\end{tabular}
		\caption{Case agreement corrections on adjectives}\label{tab:adj-case}
	\end{table*}
	
	\begin{table*}[ht]
		\begin{tabular}{llll}
			\hline
			& \textbf{Fem} & \textbf{Masc} & \textbf{Neut} \\
			\textbf{Fem} & 0 & 35 & 19 \\
			\textbf{Masc} & 39 & 0 & 28 \\
			\textbf{Neut} & 28 & 18 & 0 \\ \hline
		\end{tabular}
		\caption{Gender agreement corrections on adjectives}\label{tab:adj-gender}
	\end{table*}
	
	\begin{table*}[ht]
		\begin{tabular}{lll}
			\hline
			& \textbf{Plur} & \textbf{Sing} \\
			\textbf{Plur} & 0 & 50 \\
			\textbf{Sing} & 86 & 0 \\ \hline
		\end{tabular}
		\caption{Gender agreement corrections on adjectives}\label{tab:adj-number}
	\end{table*}
	
\end{document}


\title{Classifying Syntactic Errors in Learner Language --\\ Supplementary Material}
\maketitle
\label{sec:supplemental}

\section{Classifying SEs across Learner Levels} \label{ap:levels}

The W\&I dataset includes both corrections of texts by native speakers and by different levels of non-native speakers, we use that to analyze changes across proficiency levels. We extract the distributions of SE types from the train set for each proficiency level A--C, where A is the lowest, and for the native speakers' text. 
We compare across levels what percentage of words of a given POS tag are SEs (i.e., changed to a different POS in the correction). Results are presented in Table~\ref{ap_tab:levels_comp}; full confusion matrices are given in Appendix \S\ref{sec:wi_matrices}.

As one might expect, the more proficient the learners, the fewer SEs they make. This holds for learners across all POS types. 
Native speakers generally make fewer errors than advanced learners, but the trend is mixed. Surprisingly, native speakers rarely make SEs in POS tags that are the most error-prone for learners (top of Table \ref{ap_tab:levels_comp}). Conversely, native speakers generally do not do better than advanced learners in the least error-prone POS tags (bottom of the table). It is important to note that as the non natives have less data, the borderline changes positive and negative changes might be considered comparable rather than significant changes to one way or another. Further study of this phenomenon is deferred to future work.




Finally, we compare the syntactic proficiency of the learner language in TLE and the A--C proficiency levels of W\&I. We compute the percentage of POS tags of each type that were altered in the correction. We find the syntactic proficiency of TLE to be not too high, in most cases between level A and B. Those results agree with the prior knowledge that TLE texts are taken from B1-B2 CEFR \citep{Little2006TheCE} levels, despite the difference in the native tongues of the learners, which adds noise to the comparison.
This result thus provides further validation for \secl. 

\begin{table}[t]
\begin{small}
\centering
\begin{tabular}{lccc|c}
\toprule
      & A     & B     & C     & Native     \\ \midrule
SCONJ & 0.804 & 0.864 & 0.923 & 0.942 \\
DET   & 0.857 & 0.907 & 0.960 & 0.971 \\
ADV   & 0.844 & 0.893 & 0.945 & 0.950 \\
ADJ   & 0.875 & 0.923 & 0.962 & 0.972 \\
ADP   & 0.891 & 0.935 & 0.969 & 0.976 \\
PART  & 0.887 & 0.924 & 0.963 & 0.985 \\
AUX   & 0.901 & 0.943 & 0.973 & 0.987 \\
PROPN & 0.902 & 0.930 & 0.966 & 0.968 \\
NUM   & 0.897 & 0.929 & 0.960 & 0.950 \\
PRON  & 0.908 & 0.930 & 0.963 & 0.953 \\
NOUN  & 0.934 & 0.963 & 0.983 & 0.983 \\
CCONJ & 0.922 & 0.944 & 0.968 & 0.971 \\
VERB  & 0.945 & 0.964 & 0.983 & 0.980 \\
PUNCT & 0.978 & 0.980 & 0.990 & 0.981 \\
\bottomrule
\end{tabular}%

\caption{Percentage of unchanged POS tags per type (rows) and proficiency level (columns) in the W\&I dataset. Proficiency levels are A-C where C is the most proficient, the last column is for native speakers. Sorted by the average of columns A-C.}
\label{ap_tab:levels_comp}
\end{small}
\vspace{-0.3cm}
\end{table}

\FloatBarrier
\section{Examples of Prevalent Types of SEs Involving Replacements}\label{ssec:replacement_examples}

\subsection{POS Changes\footnote{When sentences contained other errors the corrected tokens were used for clarity of presentation.}}
\FloatBarrier

\paragraph{NOUN~\ra~VERB}
\begin{itemize}
    \item \textit{When we \sout{entrance} entered the place our problems began}.
    \item \textit{I mean it\sout{'s conflict} conflicts with your plan, what a pity!}
\end{itemize}

\paragraph{NOUN~\ra~ADJ}

\begin{itemize}
    \item \textit{These things are very \sout{convenience} convenient; on the other hand, there are lots of disadvantage points}.
    \item \textit{In this way the \sout{environment} environmental pollution will increase more and more.}
\end{itemize}

\paragraph{ADJ~\ra~NOUN}

\begin{itemize}
    \item \textit{So while I am at the Camp I like to take \sout{many} a lot of photographs and climb a mountain}.
    \item \textit{He has known research in the \sout{deeper} depths of his mind.}
\end{itemize}

\paragraph{VERB~\ra~NOUN}

\begin{itemize}
    \item \textit{I think that I am a very fast \sout{swimming} swimmer but I don't have good style and technique.}
    \item \textit{Obviously, there is no point in saying ``Famous people have a right to their own \sout{live} lives.''}
\end{itemize}

\paragraph{VERB~\ra~ADJ}

\begin{itemize}
    \item \textit{There was only a small sign \sout{stamped} stuck on the door, saying that it was ``closed for repairs''.}
    \item \textit{Sometimes you \sout{surprise} are surprised when you check the balance of your bank account.}
\end{itemize}

\paragraph{DET~\ra~PRON}

\begin{itemize}
    \item \textit{I have a camera, \sout{which} it is called Minolta X300.}
    \item \textit{To finish, I wonder if I need to bring some money or if \sout{all} everything has been already paid for.}
\end{itemize}

\paragraph{PRON~\ra~DET}

\begin{itemize}
    \item \textit{I was pleased to receive \sout{you} your letter recently.}
    \item \textit{I can say that the greatest ever invention is the invention of computers, \sout{that} which has affected both individuals and a society as a whole.}
\end{itemize}

\subsubsection{Edge-type Changes\footnote{In cases of clause-linkage errors, words participating in the relations are underlined; words inducing the relations are in boldface.}}

\paragraph{compound~\ra~amod}

\begin{itemize}
    \item \textit{Shopping is relaxing, above all on Saturdays when you have finished a \sout{work} working week and you are expecting a wonderful Sunday.}
    \item \textit{Secondly, turning to the end-of-conference party, it is planned to be in Wimbledon Common -- nice piece of \sout{England} English nature.}
\end{itemize}

\paragraph{cop~\ra~aux}

\begin{itemize}
    \item \textit{From my point of view I think he was such a very bad actor that it \sout{was} would be impossible to fine another with the same characteristics.}
    \item \textit{It should \sout{be} have been a perfect evening out, but it was the opposite.}
\end{itemize}

\paragraph{xcomp~\ra~ccomp}

\begin{itemize}
    \item \textit{It \textbf{made} people only \underline{spend} pocket-money} \ra \textit{It \textbf{meant} people only \underline{spent} pocket-money}.
    \item \textit{I went in London for a week of ``pure holiday'' thinking
\textbf{to} \underline{have} some fun coming to see your musical show} \ra \textit{I went to London for a week of ``pure holiday'', thinking I would \underline{have} some fun going to see your musical show.}
\end{itemize}

\paragraph{obl~\ra~obj}

\begin{itemize}
    \item \textit{I'm not so good in tennis but I would like to play it \textbf{for} \underline{practice}} \ra \textit{I'm not so good at tennis but I would like to play it to \textbf{get} some \underline{practice}.}
    \item \textit{However, we have to consider \sout{\textbf{about}} our human \underline{nature}.}
\end{itemize}

\paragraph{obl~\ra~advmod}

\begin{itemize}
    \item \textit{Then all the girls became jealous and stopped \sout{to laugh} laughing.}
    \item \textit{I never had a chance \sout{for staying} to stay in a tent.}
\end{itemize}

\paragraph{det~\ra~nmod:poss}

\begin{itemize}
    \item \textit{It is simple to find the way from \sout{this} your hotel to the conference.}
    \item \textit{It allowed me to worry about other things in \sout{the} my life.}
\end{itemize}

\paragraph{advmod~\ra~obl}

\begin{itemize}
    \item \textit{And it is even more difficult to predict \sout{how} what the clothes of the future will look like.}
    \item \textit{My work was for the singers, so I could meet singers and talk \sout{together} with them.}
\end{itemize}

\FloatBarrier
\section{Confusion Matrices}

\subsection{POS and edge-labels changes in TLE}\label{sec:matrices_TLE}
\FloatBarrier
Distribution of edit types in terms of POS tags and edge labels are given in Tables \ref{table:pos_confusion_matrix} and \ref{table:edge_type_confusion_matrix} respectively. Automatic ones are given in \ref{table:auto_pos_type_confusion_matrix} and \ref{table:auto_edge_type_confusion_matrix}.

\begin{table*}[ht]
\centering
\resizebox{\textwidth}{!}{%
\begin{tabular}{@{}lllllllllllllll@{}}
\toprule
 & ADJ & ADP & ADV & AUX & CCONJ & DET & NOUN & NUM & PART & PRON & PROPN & PUNCT & SCONJ & VERB \\ \midrule
ADJ & 444 & 3 & 27 & 1 & 0 & 16 & 49 & 2 & 0 & 2 & 0 & 0 & 0 & 21 \\
ADP & 4 & 727 & 11 & 0 & 2 & 10 & 5 & 0 & 1 & 0 & 0 & 0 & 6 & 23 \\
ADV & 23 & 6 & 270 & 14 & 4 & 2 & 22 & 0 & 1 & 23 & 1 & 8 & 11 & 11 \\
AUX & 4 & 1 & 7 & 372 & 0 & 0 & 8 & 0 & 8 & 15 & 0 & 0 & 0 & 21 \\
CCONJ & 1 & 4 & 3 & 0 & 40 & 1 & 0 & 0 & 0 & 0 & 0 & 1 & 5 & 0 \\
DET & 18 & 11 & 7 & 1 & 0 & 315 & 3 & 1 & 6 & 34 & 0 & 2 & 5 & 5 \\
NOUN & 50 & 5 & 22 & 5 & 0 & 3 & 1621 & 3 & 0 & 14 & 11 & 0 & 3 & 51 \\
NUM & 1 & 0 & 0 & 0 & 0 & 6 & 3 & 21 & 0 & 0 & 0 & 0 & 0 & 0 \\
PART & 0 & 0 & 4 & 4 & 0 & 6 & 0 & 0 & 2 & 0 & 0 & 0 & 2 & 1 \\
PRON & 1 & 2 & 6 & 10 & 0 & 32 & 15 & 5 & 0 & 240 & 1 & 6 & 7 & 5 \\
PROPN & 3 & 0 & 0 & 0 & 0 & 0 & 10 & 0 & 0 & 1 & 48 & 0 & 0 & 0 \\
PUNCT & 0 & 1 & 1 & 0 & 3 & 0 & 0 & 0 & 0 & 0 & 0 & 256 & 1 & 1 \\
SCONJ & 1 & 5 & 12 & 2 & 6 & 1 & 0 & 0 & 5 & 3 & 0 & 1 & 64 & 2 \\
VERB & 37 & 7 & 4 & 24 & 0 & 0 & 46 & 0 & 0 & 2 & 2 & 0 & 0 & 1818 \\ \bottomrule
\end{tabular}
}
\caption{\centering POS confusion matrix for replacement edits in absolute counts for TLE dataset with manual UD.}
\label{table:pos_confusion_matrix}
\end{table*}

\begin{table*}[ht]
\centering
\resizebox{\textwidth}{!}{%
\begin{tabular}{@{}lllllllllllllllllllllllllllllll@{}}
\toprule
 & acl & acl:relcl & advcl & advmod & amod & appos & aux & aux:pass & case & cc & ccomp & compound & conj & cop & csubj & det & expl & mark & nmod & nmod:poss & nsubj & nsubj:pass & nummod & obj & obl & obl:tmod & parataxis & punct & root & xcomp \\ \midrule
acl & 58 & 6 & 4 & 0 & 1 & 0 & 0 & 0 & 4 & 0 & 0 & 0 & 0 & 0 & 0 & 0 & 0 & 0 & 3 & 0 & 0 & 0 & 0 & 0 & 0 & 0 & 0 & 0 & 0 & 4 \\
acl:relcl & 1 & 140 & 3 & 1 & 0 & 0 & 1 & 0 & 0 & 0 & 0 & 0 & 0 & 2 & 0 & 0 & 0 & 0 & 0 & 0 & 1 & 0 & 0 & 0 & 0 & 0 & 0 & 0 & 1 & 0 \\
advcl & 7 & 3 & 321 & 3 & 0 & 0 & 4 & 0 & 3 & 0 & 3 & 0 & 2 & 2 & 1 & 0 & 0 & 0 & 0 & 0 & 1 & 0 & 0 & 1 & 11 & 0 & 4 & 0 & 4 & 17 \\
advmod & 2 & 0 & 3 & 270 & 6 & 0 & 13 & 1 & 4 & 5 & 3 & 0 & 1 & 0 & 1 & 8 & 1 & 8 & 1 & 0 & 10 & 1 & 0 & 7 & 24 & 0 & 0 & 6 & 3 & 7 \\
amod & 0 & 0 & 0 & 5 & 214 & 0 & 0 & 0 & 2 & 0 & 1 & 5 & 4 & 0 & 0 & 15 & 0 & 0 & 1 & 1 & 2 & 0 & 2 & 7 & 3 & 0 & 0 & 0 & 0 & 1 \\
appos & 0 & 0 & 0 & 0 & 0 & 14 & 0 & 0 & 0 & 0 & 0 & 1 & 0 & 0 & 0 & 0 & 0 & 0 & 0 & 0 & 0 & 0 & 0 & 0 & 0 & 0 & 1 & 0 & 0 & 0 \\
aux & 0 & 1 & 1 & 2 & 1 & 0 & 223 & 3 & 0 & 0 & 1 & 0 & 3 & 23 & 0 & 0 & 1 & 3 & 0 & 0 & 14 & 1 & 0 & 0 & 0 & 0 & 0 & 0 & 4 & 0 \\
\bottomrule
\end{tabular}
}
\caption{\centering Edge-type confusion matrix for replacement edits in absolute counts for TLE dataset with manual UD.}
\label{table:edge_type_confusion_matrix}
\end{table*}

\begin{table*}[htb] \resizebox{\textwidth}{!}{%
\begin{tabular}{lllllllllllllllllll}
 & ADJ & ADP & ADV & AUX & CCONJ & DET & INTJ & NOUN & NUM & PART & PRON & PROPN & PUNCT & SCONJ & SYM & VERB & X & Other \\
ADJ & 3907 & 24 & 61 & 2 & 2 & 192 & 1 & 78 & 4 & 0 & 15 & 4 & 11 & 4 & 0 & 48 & 0 & 0 \\
ADP & 24 & 5998 & 69 & 14 & 11 & 24 & 2 & 63 & 5 & 26 & 39 & 2 & 31 & 82 & 0 & 232 & 0 & 0 \\
ADV & 69 & 67 & 3952 & 41 & 10 & 48 & 4 & 44 & 0 & 16 & 58 & 17 & 37 & 29 & 0 & 52 & 0 & 0 \\
AUX & 3 & 10 & 27 & 5046 & 4 & 7 & 0 & 16 & 0 & 4 & 72 & 4 & 12 & 10 & 0 & 143 & 0 & 0 \\
CCONJ & 2 & 8 & 8 & 0 & 2428 & 3 & 0 & 12 & 0 & 0 & 7 & 2 & 14 & 6 & 0 & 4 & 0 & 0 \\
DET & 33 & 172 & 39 & 26 & 18 & 5451 & 0 & 33 & 6 & 10 & 78 & 0 & 19 & 30 & 0 & 76 & 0 & 0 \\
INTJ & 2 & 2 & 2 & 0 & 0 & 1 & 50 & 3 & 0 & 1 & 1 & 0 & 2 & 0 & 0 & 0 & 0 & 0 \\
NOUN & 117 & 20 & 33 & 9 & 5 & 73 & 0 & 11288 & 8 & 6 & 19 & 37 & 17 & 8 & 0 & 146 & 0 & 0 \\
NUM & 17 & 7 & 2 & 2 & 0 & 11 & 0 & 7 & 573 & 0 & 2 & 1 & 1 & 0 & 0 & 3 & 0 & 0 \\
PART & 0 & 24 & 9 & 10 & 1 & 8 & 0 & 4 & 0 & 1971 & 14 & 0 & 1 & 22 & 0 & 23 & 0 & 0 \\
PRON & 7 & 46 & 28 & 23 & 10 & 54 & 0 & 23 & 6 & 0 & 9647 & 2 & 132 & 51 & 0 & 45 & 0 & 0 \\
PROPN & 5 & 3 & 2 & 0 & 0 & 36 & 2 & 44 & 0 & 0 & 1 & 1284 & 4 & 0 & 0 & 5 & 0 & 0 \\
PUNCT & 9 & 8 & 10 & 8 & 11 & 6 & 0 & 57 & 0 & 1 & 18 & 3 & 1731 & 7 & 1 & 22 & 1 & 0 \\
SCONJ & 7 & 75 & 28 & 3 & 11 & 9 & 0 & 15 & 0 & 13 & 51 & 1 & 24 & 1846 & 0 & 28 & 1 & 0 \\
SYM & 0 & 0 & 0 & 0 & 0 & 0 & 0 & 0 & 0 & 0 & 0 & 0 & 0 & 0 & 24 & 0 & 0 & 0 \\
VERB & 80 & 74 & 47 & 153 & 2 & 22 & 2 & 127 & 2 & 21 & 19 & 6 & 18 & 4 & 0 & 9004 & 2 & 0 \\
X & 1 & 1 & 0 & 0 & 0 & 3 & 0 & 3 & 0 & 0 & 0 & 1 & 0 & 0 & 0 & 1 & 15 & 0
\end{tabular}
}
\caption{\centering POS-type confusion matrix for replacement edits in absolute counts for TLE dataset with automatic UD.}
\label{table:auto_pos_type_confusion_matrix}
\end{table*}

\begin{table*}[htb] \resizebox{\textwidth}{!}{%
\begin{tabular}{lllllllllllllllllllllllllllllllllllllllllllll}
 & acl & acl:relcl & advcl & advmod & amod & appos & aux & aux:pass & case & cc & cc:preconj & ccomp & compound & compound:prt & conj & cop & csubj & det & det:predet & discourse & expl & fixed & flat & goeswith & iobj & list & mark & nmod & nmod:npmod & nmod:poss & nmod:tmod & nsubj & nsubj:pass & nummod & obj & obl & obl:npmod & obl:tmod & parataxis & punct & reparandum & vocative & xcomp & Other \\
acl & 292 & 4 & 5 & 0 & 2 & 0 & 1 & 0 & 0 & 1 & 0 & 4 & 0 & 0 & 1 & 0 & 3 & 0 & 0 & 0 & 0 & 0 & 0 & 0 & 0 & 0 & 0 & 6 & 0 & 0 & 0 & 16 & 1 & 0 & 2 & 0 & 0 & 0 & 2 & 2 & 0 & 0 & 0 & 82 \\
acl:relcl & 10 & 755 & 8 & 2 & 0 & 0 & 0 & 0 & 0 & 0 & 0 & 2 & 1 & 0 & 3 & 2 & 0 & 0 & 0 & 1 & 0 & 0 & 0 & 0 & 0 & 0 & 5 & 3 & 0 & 0 & 0 & 8 & 2 & 0 & 12 & 18 & 1 & 10 & 7 & 0 & 0 & 0 & 3 & 153 \\
advcl & 6 & 6 & 1354 & 8 & 2 & 0 & 3 & 0 & 0 & 0 & 0 & 27 & 0 & 0 & 16 & 2 & 11 & 3 & 0 & 0 & 0 & 0 & 0 & 0 & 0 & 0 & 2 & 1 & 0 & 0 & 0 & 3 & 1 & 0 & 4 & 15 & 0 & 1 & 11 & 5 & 0 & 1 & 31 & 391 \\
advmod & 1 & 5 & 12 & 3481 & 21 & 2 & 21 & 4 & 29 & 7 & 0 & 8 & 4 & 4 & 8 & 20 & 1 & 17 & 1 & 3 & 1 & 3 & 0 & 1 & 0 & 0 & 24 & 5 & 0 & 2 & 0 & 30 & 4 & 0 & 35 & 41 & 0 & 2 & 3 & 7 & 0 & 0 & 13 & 475 \\
amod & 2 & 1 & 3 & 22 & 2345 & 0 & 0 & 0 & 23 & 1 & 1 & 0 & 22 & 0 & 3 & 1 & 0 & 174 & 3 & 0 & 0 & 0 & 0 & 0 & 0 & 0 & 1 & 2 & 0 & 4 & 0 & 10 & 0 & 4 & 21 & 9 & 2 & 1 & 1 & 4 & 0 & 0 & 1 & 121 \\
appos & 0 & 0 & 0 & 1 & 1 & 30 & 0 & 0 & 0 & 0 & 0 & 0 & 2 & 0 & 7 & 0 & 0 & 1 & 0 & 0 & 0 & 0 & 0 & 0 & 0 & 0 & 0 & 1 & 1 & 0 & 0 & 2 & 0 & 0 & 3 & 0 & 0 & 0 & 2 & 0 & 0 & 0 & 0 & 23 \\
aux & 2 & 3 & 4 & 16 & 1 & 0 & 2604 & 17 & 1 & 3 & 0 & 1 & 2 & 0 & 1 & 19 & 0 & 0 & 0 & 0 & 1 & 0 & 0 & 0 & 0 & 0 & 4 & 0 & 0 & 0 & 0 & 39 & 3 & 0 & 12 & 2 & 1 & 0 & 0 & 7 & 0 & 0 & 2 & 110 \\
\end{tabular}
}
\caption{\centering Edge-type confusion matrix for replacement edits in absolute counts for TLE dataset with automatic UD.}
\label{table:auto_edge_type_confusion_matrix}
\end{table*}

\FloatBarrier
\subsection{Output Matrices} 
\label{sec:out_matrices}
\FloatBarrier

We supply the reader with matrices of the systems of the BEA2019 shared task that shared their outputs on dev set (avoiding over analysing the test set).

\begin{table*}[htb]
\resizebox{\textwidth}{!}{%
\begin{tabular}{llllll|ll}
 & UEDIN-MS & KAKAO\&BRAIN & SHUYAO & CAMB\_CUED & TOHOKU & GOLD & Average \\
INTJ & 0.86 & 0.86 & 0.71 & 1.00 & 0.43 & 472 & 0.77 \\
PRON & 0.63 & 0.74 & 0.70 & 0.75 & 0.44 & 233 & 0.65 \\
PROPN & 0.54 & 0.44 & 0.96 & 0.60 & 0.64 & 405 & 0.64 \\
ADJ & 0.61 & 0.50 & 0.63 & 0.66 & 0.45 & 7 & 0.57 \\
DET & 0.61 & 0.40 & 0.56 & 0.68 & 0.52 & 166 & 0.55 \\
PART & 0.58 & 0.45 & 0.60 & 0.64 & 0.43 & 158 & 0.54 \\
ADV & 0.58 & 0.51 & 0.57 & 0.66 & 0.33 & 114 & 0.53 \\
AUX & 0.49 & 0.37 & 0.53 & 0.60 & 0.46 & 462 & 0.49 \\
NOUN & 0.56 & 0.35 & 0.53 & 0.53 & 0.44 & 310 & 0.48 \\
CCONJ & 0.45 & 0.61 & 0.42 & 0.59 & 0.24 & 568 & 0.46 \\
SCONJ & 0.49 & 0.48 & 0.49 & 0.54 & 0.29 & 48 & 0.46 \\
VERB & 0.50 & 0.36 & 0.50 & 0.53 & 0.37 & 191 & 0.45 \\
ADP & 0.50 & 0.36 & 0.46 & 0.52 & 0.39 & 584 & 0.45 \\
NUM & 0.46 & 0.48 & 0.42 & 0.50 & 0.31 & 618 & 0.43 \\
SYM & 0.67 & 0.00 & 0.42 & 0.42 & 0.67 & 12 & 0.43 \\
PUNCT & 0.61 & 0.21 & 0.40 & 0.40 & 0.38 & 461 & 0.40
\end{tabular}%
}
\caption{Upper bound on recall per system. The gold amount which is considered the 100\% per row and the average per line are reported on the two rightmost columns. The order of the table is descending on the average.\label{tab:average_out_pos}}
\end{table*}

\begin{table*}[htb]
\resizebox{\textwidth}{!}{%
\begin{tabular}{lllllll}
      & UEDIN-MS & KAKAO\&BRAIN & SHUYAO & CAMB-CUED & TOHOKU & GOLD \\
ADJ   & 283      & 231          & 292    & 305       & 210    & 462  \\
ADP   & 232      & 166          & 214    & 239       & 181    & 461  \\
ADV   & 273      & 242          & 268    & 313       & 155    & 472  \\
AUX   & 153      & 115          & 165    & 187       & 142    & 310  \\
CCONJ & 71       & 96           & 67     & 94        & 38     & 158  \\
DET   & 348      & 229          & 316    & 387       & 294    & 568  \\
INTJ  & 6        & 6            & 5      & 7         & 3      & 7    \\
NOUN  & 346      & 219          & 328    & 330       & 273    & 618  \\
NUM   & 22       & 23           & 20     & 24        & 15     & 48   \\
PART  & 97       & 74           & 99     & 107       & 71     & 166  \\
PRON  & 367      & 433          & 409    & 437       & 258    & 584  \\
PROPN & 62       & 50           & 109    & 68        & 73     & 114  \\
PUNCT & 116      & 40           & 76     & 77        & 73     & 191  \\
SCONJ & 114      & 111          & 115    & 126       & 67     & 233  \\
SYM   & 8        & 0            & 5      & 5         & 8      & 12   \\
VERB  & 202      & 145          & 201    & 216       & 149    & 405  \\
X     & 13       & 9            & 13     & 15        & 16     & 18  
\end{tabular}%
}
\caption{ Amount of SEs per source POS for each of the outputs and the gold standard.
\label{tab:pos-outputs}}
\end{table*}

\begin{table*}[htb] \resizebox{\textwidth}{!}{%
\begin{tabular}{lllllllllllllllllll}
 & ADJ & ADP & ADV & AUX & CCONJ & DET & INTJ & NOUN & NUM & PART & PRON & PROPN & PUNCT & SCONJ & SYM & VERB & X & Other \\
ADJ & 6281 & 20 & 21 & 4 & 3 & 109 & 0 & 46 & 0 & 0 & 9 & 8 & 28 & 2 & 0 & 33 & 0 & 0 \\
ADP & 17 & 7649 & 10 & 10 & 0 & 3 & 0 & 25 & 1 & 22 & 16 & 1 & 52 & 35 & 0 & 40 & 0 & 0 \\
ADV & 28 & 18 & 4650 & 23 & 5 & 17 & 8 & 17 & 0 & 3 & 25 & 2 & 95 & 14 & 0 & 18 & 0 & 0 \\
AUX & 2 & 1 & 8 & 5680 & 4 & 2 & 0 & 19 & 0 & 10 & 51 & 2 & 11 & 2 & 0 & 41 & 0 & 0 \\
CCONJ & 0 & 2 & 2 & 1 & 2946 & 2 & 0 & 9 & 0 & 0 & 2 & 1 & 51 & 0 & 0 & 1 & 0 & 0 \\
DET & 15 & 112 & 13 & 13 & 8 & 7175 & 1 & 11 & 1 & 8 & 31 & 2 & 61 & 13 & 0 & 58 & 1 & 0 \\
INTJ & 0 & 1 & 0 & 0 & 0 & 0 & 92 & 1 & 0 & 0 & 0 & 1 & 2 & 0 & 0 & 1 & 0 & 0 \\
NOUN & 67 & 7 & 19 & 10 & 5 & 35 & 0 & 16050 & 2 & 3 & 18 & 33 & 55 & 5 & 0 & 87 & 0 & 0 \\
NUM & 3 & 1 & 2 & 1 & 0 & 3 & 0 & 6 & 628 & 0 & 0 & 0 & 4 & 1 & 0 & 1 & 0 & 0 \\
PART & 2 & 10 & 6 & 9 & 1 & 2 & 0 & 9 & 0 & 2523 & 13 & 1 & 6 & 20 & 0 & 18 & 0 & 0 \\
PRON & 1 & 13 & 16 & 17 & 6 & 12 & 0 & 25 & 0 & 2 & 8016 & 1 & 235 & 14 & 0 & 25 & 0 & 0 \\
PROPN & 14 & 1 & 0 & 0 & 1 & 16 & 0 & 19 & 0 & 1 & 1 & 1926 & 6 & 0 & 0 & 3 & 0 & 0 \\
PUNCT & 7 & 8 & 10 & 8 & 8 & 5 & 1 & 29 & 1 & 6 & 5 & 6 & 8679 & 11 & 0 & 11 & 0 & 0 \\
SCONJ & 0 & 25 & 7 & 0 & 1 & 2 & 0 & 6 & 0 & 17 & 12 & 1 & 31 & 1882 & 0 & 12 & 0 & 0 \\
SYM & 1 & 0 & 0 & 0 & 0 & 0 & 0 & 1 & 0 & 1 & 1 & 0 & 3 & 0 & 45 & 1 & 0 & 0 \\
VERB & 21 & 9 & 9 & 44 & 3 & 7 & 1 & 67 & 0 & 5 & 7 & 5 & 20 & 3 & 0 & 9973 & 1 & 0 \\
X & 0 & 0 & 2 & 0 & 0 & 2 & 0 & 2 & 0 & 0 & 2 & 0 & 2 & 0 & 0 & 3 & 65 & 0
\end{tabular}
}
\caption{UEDIN-MS POS confusion matrix on W\&I development.}
\end{table*}

\begin{table*}[htb] \resizebox{\textwidth}{!}{%
\begin{tabular}{lllllllllllllllllll}
 & ADJ & ADP & ADV & AUX & CCONJ & DET & INTJ & NOUN & NUM & PART & PRON & PROPN & PUNCT & SCONJ & SYM & VERB & X & Other \\
ADJ & 6354 & 16 & 18 & 3 & 1 & 92 & 0 & 37 & 0 & 1 & 2 & 6 & 13 & 0 & 0 & 21 & 0 & 0 \\
ADP & 19 & 7700 & 14 & 6 & 0 & 4 & 0 & 30 & 1 & 12 & 8 & 4 & 11 & 25 & 0 & 47 & 0 & 0 \\
ADV & 15 & 12 & 4768 & 19 & 4 & 13 & 4 & 15 & 0 & 2 & 12 & 3 & 25 & 11 & 0 & 20 & 0 & 0 \\
AUX & 1 & 2 & 10 & 5691 & 3 & 1 & 0 & 22 & 0 & 12 & 41 & 0 & 6 & 1 & 0 & 43 & 0 & 0 \\
CCONJ & 2 & 0 & 1 & 1 & 2979 & 2 & 0 & 6 & 0 & 1 & 2 & 1 & 16 & 0 & 1 & 5 & 0 & 0 \\
DET & 13 & 93 & 21 & 10 & 9 & 7229 & 0 & 12 & 0 & 5 & 17 & 1 & 45 & 10 & 0 & 57 & 1 & 0 \\
INTJ & 0 & 0 & 0 & 0 & 0 & 0 & 95 & 0 & 0 & 0 & 2 & 0 & 1 & 0 & 0 & 0 & 0 & 0 \\
NOUN & 46 & 6 & 20 & 2 & 5 & 33 & 0 & 16123 & 3 & 3 & 22 & 21 & 31 & 5 & 0 & 76 & 0 & 0 \\
NUM & 1 & 0 & 0 & 0 & 1 & 3 & 0 & 5 & 635 & 0 & 1 & 0 & 1 & 1 & 0 & 2 & 0 & 0 \\
PART & 2 & 7 & 4 & 9 & 0 & 0 & 0 & 11 & 0 & 2549 & 11 & 1 & 2 & 13 & 0 & 11 & 0 & 0 \\
PRON & 2 & 11 & 13 & 14 & 7 & 10 & 0 & 28 & 0 & 2 & 8125 & 2 & 128 & 13 & 0 & 28 & 0 & 0 \\
PROPN & 19 & 5 & 0 & 1 & 0 & 13 & 0 & 24 & 0 & 0 & 3 & 1915 & 4 & 0 & 0 & 3 & 1 & 0 \\
PUNCT & 11 & 2 & 8 & 1 & 5 & 2 & 1 & 22 & 0 & 0 & 4 & 2 & 8722 & 4 & 1 & 10 & 0 & 0 \\
SCONJ & 0 & 12 & 9 & 4 & 1 & 3 & 0 & 5 & 0 & 9 & 7 & 0 & 8 & 1929 & 0 & 9 & 0 & 0 \\
SYM & 2 & 0 & 0 & 0 & 0 & 0 & 0 & 2 & 0 & 0 & 0 & 0 & 3 & 0 & 45 & 1 & 0 & 0 \\
VERB & 17 & 7 & 7 & 30 & 3 & 6 & 3 & 44 & 0 & 7 & 9 & 2 & 11 & 2 & 0 & 10026 & 1 & 0 \\
X & 1 & 0 & 2 & 1 & 0 & 1 & 0 & 6 & 0 & 0 & 1 & 2 & 1 & 0 & 0 & 1 & 62 & 0
\end{tabular}
}
\caption{AIP-TOHOKU POS confusion matrix on W\&I development.}
\end{table*}

\begin{table*}[htb] \resizebox{\textwidth}{!}{%
\begin{tabular}{lllllllllllllllllll}
 & ADJ & ADP & ADV & AUX & CCONJ & DET & INTJ & NOUN & NUM & PART & PRON & PROPN & PUNCT & SCONJ & SYM & VERB & X & Other \\
ADJ & 6333 & 9 & 16 & 2 & 0 & 98 & 0 & 41 & 0 & 0 & 5 & 7 & 26 & 0 & 0 & 27 & 0 & 0 \\
ADP & 10 & 7715 & 7 & 6 & 1 & 4 & 0 & 14 & 0 & 10 & 8 & 1 & 39 & 33 & 0 & 33 & 0 & 0 \\
ADV & 20 & 16 & 4681 & 13 & 1 & 12 & 10 & 6 & 0 & 2 & 12 & 4 & 124 & 9 & 0 & 13 & 0 & 0 \\
AUX & 0 & 0 & 2 & 5718 & 1 & 1 & 0 & 11 & 0 & 9 & 32 & 0 & 11 & 4 & 0 & 44 & 0 & 0 \\
CCONJ & 0 & 0 & 2 & 0 & 2921 & 2 & 0 & 3 & 0 & 0 & 0 & 2 & 86 & 0 & 0 & 1 & 0 & 0 \\
DET & 12 & 59 & 4 & 2 & 3 & 7294 & 1 & 6 & 0 & 5 & 21 & 0 & 81 & 10 & 0 & 25 & 0 & 0 \\
INTJ & 0 & 0 & 0 & 0 & 0 & 0 & 92 & 0 & 0 & 0 & 2 & 1 & 2 & 0 & 0 & 1 & 0 & 0 \\
NOUN & 31 & 3 & 11 & 2 & 3 & 23 & 0 & 16177 & 3 & 0 & 11 & 17 & 48 & 4 & 0 & 63 & 0 & 0 \\
NUM & 3 & 1 & 0 & 1 & 0 & 4 & 0 & 5 & 627 & 0 & 0 & 0 & 9 & 0 & 0 & 0 & 0 & 0 \\
PART & 1 & 9 & 3 & 13 & 0 & 0 & 0 & 8 & 0 & 2546 & 8 & 0 & 4 & 18 & 0 & 10 & 0 & 0 \\
PRON & 2 & 12 & 6 & 12 & 0 & 8 & 0 & 17 & 0 & 1 & 7950 & 1 & 346 & 11 & 0 & 17 & 0 & 0 \\
PROPN & 2 & 1 & 0 & 0 & 0 & 12 & 0 & 20 & 0 & 0 & 1 & 1938 & 13 & 0 & 0 & 1 & 0 & 0 \\
PUNCT & 1 & 6 & 3 & 2 & 4 & 0 & 0 & 10 & 0 & 1 & 3 & 0 & 8755 & 5 & 0 & 5 & 0 & 0 \\
SCONJ & 0 & 24 & 7 & 0 & 0 & 1 & 0 & 4 & 0 & 18 & 10 & 0 & 37 & 1885 & 0 & 10 & 0 & 0 \\
SYM & 0 & 0 & 0 & 0 & 0 & 0 & 0 & 0 & 0 & 0 & 0 & 0 & 0 & 0 & 53 & 0 & 0 & 0 \\
VERB & 16 & 4 & 5 & 27 & 1 & 10 & 1 & 47 & 0 & 1 & 5 & 3 & 21 & 4 & 0 & 10030 & 0 & 0 \\
X & 0 & 0 & 1 & 1 & 0 & 1 & 0 & 4 & 0 & 0 & 1 & 0 & 0 & 0 & 0 & 1 & 69 & 0
\end{tabular}
}
\caption{Kakao\&Brain POS confusion matrix on W\&I development.}
\end{table*}

\begin{table*}[htb] \resizebox{\textwidth}{!}{%
\begin{tabular}{lllllllllllllllllll}
 & ADJ & ADP & ADV & AUX & CCONJ & DET & INTJ & NOUN & NUM & PART & PRON & PROPN & PUNCT & SCONJ & SYM & VERB & X & Other \\
ADJ & 6272 & 20 & 30 & 1 & 0 & 100 & 0 & 57 & 3 & 0 & 5 & 9 & 33 & 0 & 0 & 34 & 0 & 0 \\
ADP & 13 & 7667 & 11 & 6 & 1 & 3 & 0 & 20 & 0 & 17 & 14 & 2 & 49 & 37 & 0 & 41 & 0 & 0 \\
ADV & 28 & 18 & 4655 & 15 & 6 & 15 & 6 & 16 & 0 & 1 & 16 & 2 & 115 & 12 & 0 & 18 & 0 & 0 \\
AUX & 2 & 2 & 10 & 5668 & 1 & 2 & 0 & 23 & 0 & 9 & 48 & 3 & 16 & 1 & 0 & 48 & 0 & 0 \\
CCONJ & 0 & 1 & 1 & 1 & 2950 & 3 & 0 & 7 & 0 & 0 & 2 & 0 & 52 & 0 & 0 & 0 & 0 & 0 \\
DET & 17 & 98 & 9 & 11 & 6 & 7207 & 2 & 7 & 0 & 6 & 25 & 1 & 81 & 9 & 0 & 44 & 0 & 0 \\
INTJ & 0 & 0 & 0 & 0 & 0 & 0 & 93 & 0 & 0 & 0 & 0 & 3 & 2 & 0 & 0 & 0 & 0 & 0 \\
NOUN & 64 & 9 & 20 & 4 & 6 & 26 & 0 & 16068 & 11 & 1 & 15 & 26 & 59 & 5 & 0 & 82 & 0 & 0 \\
NUM & 1 & 1 & 2 & 0 & 0 & 2 & 0 & 6 & 630 & 0 & 0 & 0 & 7 & 0 & 0 & 1 & 0 & 0 \\
PART & 2 & 11 & 6 & 10 & 1 & 2 & 0 & 12 & 0 & 2521 & 11 & 0 & 4 & 22 & 0 & 18 & 0 & 0 \\
PRON & 3 & 14 & 11 & 13 & 3 & 11 & 0 & 24 & 2 & 2 & 7974 & 1 & 288 & 14 & 0 & 23 & 0 & 0 \\
PROPN & 12 & 3 & 3 & 0 & 0 & 16 & 1 & 50 & 0 & 1 & 6 & 1879 & 10 & 0 & 0 & 6 & 1 & 0 \\
PUNCT & 6 & 6 & 6 & 4 & 1 & 0 & 0 & 35 & 1 & 0 & 1 & 0 & 8719 & 10 & 0 & 6 & 0 & 0 \\
SCONJ & 1 & 21 & 7 & 0 & 1 & 2 & 0 & 7 & 0 & 18 & 8 & 0 & 38 & 1881 & 0 & 12 & 0 & 0 \\
SYM & 1 & 0 & 0 & 0 & 0 & 0 & 0 & 2 & 0 & 0 & 0 & 0 & 1 & 0 & 48 & 1 & 0 & 0 \\
VERB & 17 & 11 & 10 & 34 & 1 & 7 & 1 & 77 & 1 & 2 & 7 & 4 & 26 & 3 & 0 & 9974 & 0 & 0 \\
X & 1 & 0 & 2 & 1 & 0 & 1 & 0 & 3 & 0 & 1 & 1 & 0 & 2 & 0 & 0 & 1 & 65 & 0
\end{tabular}
}
\caption{SHUYAO POS confusion matrix on W\&I development.}
\end{table*}

\begin{table*}[htb] \resizebox{\textwidth}{!}{%
\begin{tabular}{lllllllllllllllllll}
 & ADJ & ADP & ADV & AUX & CCONJ & DET & INTJ & NOUN & NUM & PART & PRON & PROPN & PUNCT & SCONJ & SYM & VERB & X & Other \\
ADJ & 6259 & 23 & 24 & 3 & 4 & 118 & 0 & 55 & 0 & 0 & 8 & 9 & 30 & 0 & 0 & 31 & 0 & 0 \\
ADP & 15 & 7642 & 8 & 12 & 1 & 5 & 0 & 35 & 0 & 21 & 12 & 1 & 61 & 26 & 0 & 42 & 0 & 0 \\
ADV & 23 & 19 & 4610 & 20 & 9 & 17 & 10 & 21 & 0 & 2 & 21 & 4 & 122 & 19 & 0 & 26 & 0 & 0 \\
AUX & 3 & 1 & 14 & 5646 & 3 & 4 & 0 & 24 & 0 & 12 & 56 & 2 & 13 & 2 & 0 & 53 & 0 & 0 \\
CCONJ & 3 & 1 & 4 & 1 & 2923 & 2 & 0 & 7 & 0 & 0 & 5 & 1 & 69 & 0 & 0 & 1 & 0 & 0 \\
DET & 17 & 122 & 16 & 12 & 10 & 7136 & 1 & 17 & 2 & 8 & 30 & 2 & 72 & 15 & 0 & 63 & 0 & 0 \\
INTJ & 0 & 0 & 0 & 0 & 0 & 0 & 91 & 0 & 0 & 0 & 2 & 1 & 3 & 0 & 0 & 1 & 0 & 0 \\
NOUN & 50 & 9 & 15 & 7 & 8 & 33 & 0 & 16066 & 1 & 5 & 21 & 32 & 60 & 3 & 0 & 86 & 0 & 0 \\
NUM & 3 & 1 & 0 & 0 & 1 & 3 & 0 & 6 & 626 & 1 & 1 & 0 & 6 & 1 & 0 & 1 & 0 & 0 \\
PART & 3 & 10 & 12 & 14 & 1 & 0 & 0 & 12 & 0 & 2513 & 9 & 0 & 3 & 24 & 0 & 19 & 0 & 0 \\
PRON & 4 & 21 & 9 & 17 & 10 & 9 & 0 & 27 & 0 & 2 & 7946 & 3 & 293 & 16 & 0 & 26 & 0 & 0 \\
PROPN & 10 & 1 & 0 & 0 & 0 & 16 & 0 & 32 & 1 & 0 & 1 & 1920 & 6 & 0 & 0 & 1 & 0 & 0 \\
PUNCT & 1 & 4 & 7 & 5 & 4 & 2 & 2 & 28 & 0 & 2 & 6 & 3 & 8718 & 7 & 0 & 6 & 0 & 0 \\
SCONJ & 1 & 20 & 8 & 1 & 1 & 2 & 0 & 6 & 0 & 13 & 12 & 0 & 51 & 1870 & 0 & 11 & 0 & 0 \\
SYM & 0 & 0 & 0 & 0 & 0 & 0 & 0 & 1 & 0 & 0 & 0 & 0 & 3 & 0 & 48 & 1 & 0 & 0 \\
VERB & 25 & 9 & 11 & 48 & 4 & 6 & 1 & 63 & 0 & 6 & 13 & 1 & 25 & 3 & 0 & 9959 & 1 & 0 \\
X & 0 & 0 & 2 & 1 & 1 & 2 & 0 & 6 & 0 & 1 & 1 & 0 & 1 & 0 & 0 & 0 & 63 & 0
\end{tabular}
}
\caption{CAMB-CUED POS confusion matrix on W\&I development.}
\end{table*}
\FloatBarrier
\subsection{W\&I by levels}\label{sec:wi_matrices}
\FloatBarrier

We provide a separate matrix for each level of learner proficiency in the W\&I corpus and the native proficiency of LOCNESS. 
\begin{table*}[htb]
\resizebox{\textwidth}{!}{%
\begin{tabular}{llllllllllllllllll}
 & ADJ & ADP & ADV & AUX & CCONJ & DET & INTJ & NOUN & NUM & PART & PRON & PROPN & PUNCT & SCONJ & SYM & VERB & X \\
ADJ & 11594 & 70 & 190 & 11 & 11 & 442 & 2 & 456 & 7 & 5 & 36 & 65 & 123 & 12 & 0 & 225 & 0 \\
ADP & 101 & 13880 & 167 & 35 & 33 & 70 & 7 & 233 & 12 & 103 & 82 & 14 & 266 & 193 & 1 & 388 & 0 \\
ADV & 181 & 118 & 8493 & 119 & 78 & 107 & 28 & 173 & 5 & 23 & 134 & 18 & 361 & 80 & 1 & 147 & 0 \\
AUX & 23 & 8 & 59 & 10907 & 9 & 16 & 1 & 111 & 1 & 39 & 324 & 14 & 75 & 8 & 1 & 513 & 0 \\
CCONJ & 13 & 17 & 9 & 4 & 5843 & 9 & 0 & 45 & 3 & 2 & 8 & 0 & 365 & 6 & 0 & 11 & 0 \\
DET & 82 & 522 & 105 & 115 & 66 & 12806 & 5 & 147 & 22 & 22 & 186 & 9 & 341 & 112 & 0 & 410 & 0 \\
INTJ & 4 & 0 & 4 & 2 & 0 & 3 & 312 & 8 & 0 & 1 & 1 & 5 & 9 & 1 & 0 & 5 & 0 \\
NOUN & 372 & 71 & 170 & 71 & 25 & 156 & 12 & 31992 & 28 & 10 & 133 & 235 & 300 & 32 & 1 & 658 & 4 \\
NUM & 8 & 17 & 8 & 7 & 1 & 32 & 0 & 48 & 1405 & 1 & 10 & 1 & 14 & 1 & 2 & 11 & 0 \\
PART & 14 & 79 & 51 & 81 & 11 & 11 & 1 & 63 & 1 & 4845 & 38 & 7 & 32 & 104 & 0 & 125 & 0 \\
PRON & 24 & 94 & 68 & 81 & 48 & 78 & 1 & 165 & 6 & 23 & 19290 & 15 & 1110 & 100 & 0 & 137 & 1 \\
PROPN & 32 & 14 & 8 & 2 & 4 & 45 & 6 & 156 & 1 & 1 & 7 & 3297 & 62 & 3 & 0 & 16 & 0 \\
PUNCT & 21 & 22 & 26 & 9 & 24 & 8 & 5 & 140 & 3 & 11 & 44 & 11 & 17928 & 26 & 1 & 58 & 1 \\
SCONJ & 16 & 113 & 70 & 7 & 20 & 15 & 1 & 46 & 1 & 61 & 107 & 9 & 254 & 3270 & 0 & 76 & 0 \\
SYM & 0 & 2 & 0 & 0 & 1 & 3 & 0 & 0 & 0 & 0 & 0 & 0 & 1 & 0 & 77 & 1 & 1 \\
VERB & 169 & 59 & 65 & 251 & 9 & 40 & 4 & 393 & 1 & 21 & 43 & 28 & 103 & 21 & 3 & 21005 & 1 \\
X & 1 & 0 & 3 & 1 & 1 & 2 & 0 & 5 & 0 & 1 & 0 & 4 & 2 & 1 & 2 & 1 & 111
\end{tabular}%
}
\caption{Matrix for POS on W\&I training set of A (lowest) level learners.}
\end{table*}
\begin{table*}[htb]
\resizebox{\textwidth}{!}{%
\begin{tabular}{llllllllllllllllll}
 & ADJ & ADP & ADV & AUX & CCONJ & DET & INTJ & NOUN & NUM & PART & PRON & PROPN & PUNCT & SCONJ & SYM & VERB & X \\
ADJ & 16232 & 66 & 155 & 14 & 9 & 385 & 2 & 319 & 10 & 2 & 37 & 64 & 121 & 17 & 0 & 158 & 1 \\
ADP & 80 & 19910 & 132 & 37 & 31 & 57 & 4 & 166 & 15 & 61 & 58 & 11 & 263 & 165 & 1 & 307 & 1 \\
ADV & 142 & 96 & 11753 & 126 & 40 & 67 & 27 & 158 & 1 & 31 & 119 & 8 & 365 & 89 & 0 & 134 & 1 \\
AUX & 13 & 6 & 61 & 14903 & 4 & 15 & 1 & 92 & 2 & 33 & 213 & 8 & 92 & 14 & 0 & 340 & 0 \\
CCONJ & 13 & 13 & 11 & 1 & 7499 & 5 & 1 & 30 & 0 & 2 & 11 & 3 & 336 & 8 & 0 & 9 & 1 \\
DET & 70 & 536 & 99 & 88 & 63 & 17932 & 6 & 109 & 20 & 14 & 131 & 6 & 281 & 96 & 0 & 316 & 0 \\
INTJ & 3 & 0 & 6 & 0 & 0 & 0 & 269 & 6 & 0 & 0 & 3 & 3 & 8 & 1 & 0 & 2 & 0 \\
NOUN & 288 & 53 & 124 & 36 & 21 & 151 & 8 & 42212 & 31 & 6 & 71 & 181 & 196 & 32 & 2 & 437 & 2 \\
NUM & 8 & 9 & 3 & 2 & 0 & 31 & 0 & 44 & 1756 & 0 & 6 & 6 & 17 & 1 & 0 & 8 & 0 \\
PART & 16 & 60 & 49 & 43 & 4 & 12 & 0 & 55 & 0 & 6880 & 42 & 8 & 35 & 114 & 0 & 130 & 0 \\
PRON & 14 & 65 & 44 & 58 & 45 & 74 & 1 & 139 & 13 & 15 & 23771 & 9 & 1101 & 100 & 0 & 109 & 0 \\
PROPN & 20 & 8 & 15 & 4 & 2 & 58 & 1 & 169 & 0 & 1 & 9 & 4854 & 59 & 5 & 0 & 11 & 1 \\
PUNCT & 48 & 20 & 30 & 18 & 35 & 21 & 2 & 172 & 0 & 8 & 34 & 14 & 23620 & 26 & 1 & 64 & 0 \\
SCONJ & 10 & 115 & 76 & 10 & 17 & 14 & 0 & 46 & 1 & 70 & 63 & 7 & 277 & 4720 & 0 & 34 & 0 \\
SYM & 1 & 0 & 0 & 0 & 0 & 0 & 0 & 3 & 0 & 0 & 0 & 0 & 0 & 0 & 79 & 1 & 0 \\
VERB & 165 & 44 & 65 & 182 & 7 & 37 & 2 & 329 & 2 & 19 & 32 & 25 & 96 & 22 & 1 & 27432 & 0 \\
X & 0 & 0 & 3 & 0 & 0 & 2 & 0 & 8 & 0 & 1 & 2 & 4 & 5 & 0 & 2 & 4 & 143
\end{tabular}%
}
\caption{Matrix for POS on W\&I training set of B (mid) level learners.}
\end{table*}

\begin{table*}[htb]
\resizebox{\textwidth}{!}{%
\begin{tabular}{llllllllllllllllll}
 & ADJ & ADP & ADV & AUX & CCONJ & DET & INTJ & NOUN & NUM & PART & PRON & PROPN & PUNCT & SCONJ & SYM & VERB & X \\
ADJ & 14938 & 27 & 81 & 8 & 3 & 174 & 0 & 139 & 2 & 3 & 13 & 23 & 53 & 1 & 0 & 67 & 0 \\
ADP & 35 & 18358 & 60 & 13 & 13 & 14 & 2 & 58 & 3 & 19 & 24 & 1 & 161 & 48 & 1 & 133 & 1 \\
ADV & 64 & 50 & 11003 & 34 & 26 & 41 & 7 & 74 & 2 & 8 & 45 & 2 & 217 & 28 & 0 & 47 & 0 \\
AUX & 8 & 8 & 22 & 12706 & 6 & 4 & 0 & 38 & 0 & 13 & 81 & 5 & 49 & 5 & 0 & 109 & 0 \\
CCONJ & 1 & 12 & 4 & 6 & 6767 & 9 & 0 & 15 & 0 & 2 & 6 & 0 & 167 & 0 & 0 & 3 & 0 \\
DET & 22 & 226 & 26 & 24 & 18 & 16736 & 0 & 32 & 2 & 4 & 49 & 5 & 142 & 19 & 0 & 125 & 1 \\
INTJ & 1 & 1 & 4 & 0 & 0 & 0 & 216 & 2 & 0 & 0 & 0 & 1 & 2 & 0 & 0 & 0 & 0 \\
NOUN & 116 & 32 & 29 & 14 & 5 & 63 & 2 & 37581 & 8 & 2 & 38 & 52 & 91 & 4 & 1 & 172 & 2 \\
NUM & 10 & 1 & 3 & 1 & 0 & 9 & 0 & 18 & 1501 & 0 & 7 & 0 & 6 & 0 & 1 & 5 & 1 \\
PART & 3 & 18 & 26 & 19 & 4 & 2 & 0 & 31 & 2 & 6167 & 21 & 5 & 18 & 57 & 0 & 29 & 0 \\
PRON & 8 & 30 & 18 & 31 & 4 & 20 & 0 & 32 & 2 & 0 & 20363 & 3 & 574 & 32 & 0 & 37 & 0 \\
PROPN & 14 & 2 & 7 & 1 & 1 & 25 & 0 & 50 & 0 & 1 & 1 & 4511 & 52 & 0 & 0 & 5 & 0 \\
PUNCT & 25 & 9 & 15 & 5 & 11 & 6 & 2 & 88 & 1 & 5 & 20 & 3 & 21344 & 7 & 0 & 14 & 0 \\
SCONJ & 8 & 59 & 34 & 1 & 5 & 10 & 1 & 19 & 0 & 20 & 32 & 5 & 144 & 4239 & 0 & 16 & 0 \\
SYM & 1 & 1 & 1 & 0 & 1 & 1 & 1 & 4 & 0 & 0 & 0 & 0 & 9 & 0 & 84 & 0 & 0 \\
VERB & 58 & 12 & 18 & 66 & 2 & 13 & 1 & 148 & 2 & 3 & 11 & 4 & 75 & 3 & 0 & 24299 & 2 \\
X & 0 & 0 & 1 & 0 & 0 & 2 & 0 & 1 & 0 & 0 & 0 & 2 & 1 & 0 & 0 & 0 & 120
\end{tabular}%
}
\caption{Matrix for POS on W\&I training set of C (highest) level learners.}
\end{table*}
\begin{table*}[htb]
\resizebox{\textwidth}{!}{%
\begin{tabular}{llllllllllllllllll}
 & ADJ & ADP & ADV & AUX & CCONJ & DET & INTJ & NOUN & NUM & PART & PRON & PROPN & PUNCT & SCONJ & SYM & VERB & X \\
ADJ & 1753 & 1 & 5 & 0 & 0 & 3 & 0 & 18 & 0 & 0 & 1 & 4 & 10 & 0 & 0 & 8 & 0 \\
ADP & 4 & 2228 & 5 & 0 & 1 & 1 & 0 & 9 & 0 & 3 & 1 & 0 & 20 & 6 & 0 & 5 & 0 \\
ADV & 6 & 6 & 1264 & 2 & 1 & 0 & 0 & 7 & 0 & 0 & 1 & 0 & 41 & 2 & 0 & 0 & 1 \\
AUX & 0 & 0 & 0 & 1720 & 0 & 0 & 0 & 4 & 0 & 1 & 2 & 0 & 11 & 0 & 0 & 5 & 0 \\
CCONJ & 0 & 2 & 0 & 0 & 763 & 0 & 0 & 2 & 0 & 0 & 0 & 0 & 18 & 1 & 0 & 0 & 0 \\
DET & 3 & 17 & 2 & 4 & 3 & 2138 & 0 & 0 & 0 & 0 & 2 & 0 & 27 & 2 & 0 & 3 & 0 \\
INTJ & 0 & 0 & 0 & 0 & 0 & 0 & 9 & 0 & 0 & 0 & 0 & 0 & 0 & 0 & 0 & 0 & 0 \\
NOUN & 10 & 2 & 7 & 0 & 0 & 2 & 0 & 4554 & 1 & 0 & 1 & 8 & 24 & 3 & 1 & 19 & 0 \\
NUM & 0 & 0 & 0 & 0 & 0 & 0 & 0 & 1 & 153 & 0 & 0 & 0 & 4 & 0 & 2 & 0 & 1 \\
PART & 0 & 2 & 0 & 1 & 0 & 0 & 0 & 1 & 0 & 643 & 2 & 0 & 4 & 0 & 0 & 0 & 0 \\
PRON & 0 & 1 & 2 & 1 & 2 & 0 & 0 & 1 & 2 & 0 & 1422 & 0 & 55 & 2 & 0 & 4 & 0 \\
PROPN & 5 & 0 & 0 & 0 & 0 & 1 & 0 & 11 & 0 & 0 & 0 & 568 & 1 & 0 & 0 & 1 & 0 \\
PUNCT & 4 & 3 & 2 & 1 & 0 & 1 & 0 & 22 & 1 & 1 & 1 & 1 & 2304 & 3 & 0 & 4 & 0 \\
SCONJ & 0 & 3 & 3 & 0 & 0 & 0 & 0 & 1 & 0 & 1 & 4 & 0 & 16 & 504 & 0 & 3 & 0 \\
SYM & 2 & 0 & 0 & 0 & 0 & 0 & 0 & 3 & 0 & 0 & 0 & 0 & 2 & 0 & 12 & 1 & 1 \\
VERB & 2 & 1 & 2 & 4 & 0 & 0 & 0 & 23 & 0 & 0 & 1 & 4 & 12 & 0 & 0 & 2459 & 0 \\
X & 0 & 0 & 0 & 0 & 0 & 1 & 0 & 1 & 0 & 0 & 0 & 0 & 2 & 0 & 1 & 0 & 19
\end{tabular}%
}
\caption{Matrix for POS on LOCNESS training set of N (native) writers.}
\end{table*}

\FloatBarrier
\subsection{Comparison to NUCLE}
Confusion matrix between SeRCL and NUCLE types over the NUCLE train set is given in \ref{tab:m2_stx}. SeRCL types with less than 30 occurrences are omitted. Aggregations are find in the rightmost columns (and do not include -rloc)
\begin{table*}[htb]
\resizebox{\linewidth}{!}{%
%
%
}
\caption{NUCLE and SerCL confusion matrix \label{tab:m2_stx}}
\end{table*}

\FloatBarrier
\subsection{Morphological feature-changes in edits}\label{sec:matrices_morph}
\subsubsection{Agreement features on adjectives}
\FloatBarrier

We provide  the morphological aware matrices, delving into finer details than the POS tag and hence allowing for inner classifications of the cases were the POS tag was kept but its characteristics were not.
\begin{table*}[ht]
%
\caption{Case agreement corrections on adjectives}\label{tab:adj-case}
\end{table*}

\begin{table*}[ht]
\begin{tabular}{llll}
\hline
 & \textbf{Fem} & \textbf{Masc} & \textbf{Neut} \\
\textbf{Fem} & 0 & 35 & 19 \\
\textbf{Masc} & 39 & 0 & 28 \\
\textbf{Neut} & 28 & 18 & 0 \\ \hline
\end{tabular}
\caption{Gender agreement corrections on adjectives}\label{tab:adj-gender}
\end{table*}

\begin{table*}[ht]
%
\caption{Gender agreement corrections on adjectives}\label{tab:adj-number}
\end{table*}
\FloatBarrier

\bibliography{emnlp-ijcnlp-2019}
\bibliographystyle{acl_natbib}